\documentclass{elsarticle}
\usepackage{amsmath}
\usepackage{amsfonts}
\usepackage{comment}
\usepackage{multirow}
\usepackage{graphicx} 
\usepackage{booktabs}
\usepackage{xcolor}
\usepackage{url}
\usepackage{adjustbox}

\renewcommand{\textcolor}[2]{#2}

\begin{document}
\begin{frontmatter}

\title{Parameter-Efficient Continual Fine-Tuning: A Survey}
\author[unipi]{Eric Nuertey Coleman} 
\ead{eric.coleman@phd.unipi.it}

\author[unipi]{Luigi Quarantiello\corref{1}}
\ead{luigi.quarantiello@phd.unipi.it}
\author[torino]{Ziyue Liu}
\ead{ziyue.liu@polito.it}
\author[auckland]{Qinwen Yang}
\ead{Yangqq7160@gmail.com}
\author[iit]{Samrat Mukherjee}
\ead{samratpisv123@gmail.com
}
\author[warwick]{Julio Hurtado}
\ead{julio.hurtado@warwick.ac.uk}
\author[luiss]{Vincenzo Lomonaco}
\ead{vlomonaco@luiss.it}
\affiliation[unipi]{organization={University of Pisa},
            addressline={Largo Bruno Pontecorvo 3},
            city={Pisa},
            postcode={56127},
            country={Italy}}
\affiliation[warwick]{organization={University of Warwick},
            addressline={Coventry CV4 7AL},
            country={United Kingdom}}
\affiliation[iit]{organization={Indian Institute of Technology},
            addressline={Main Gate Rd, IIT Area, Powai, Mumbai, Maharashtra 400076},
            country={India}}
\affiliation[torino]{organization={Politecnico di Torino},
            addressline={Corso Duca degli Abruzzi 24},
            city={Torino},
            postcode={10129},
            country={Italy}}
\affiliation[auckland]{organization={University of Auckland},
            addressline={34 Princes Street, Auckland Central, Auckland 1010},
            country={New Zealand}}
\affiliation[luiss]{organization={LUISS Guido Carli University},
            addressline={Viale Romania, 32, Rome},
            city={Rome},
            postcode={00197},
            country={Italy}}
\cortext[1]{Corresponding author}

\begin{abstract}
The advent of large-scale pre-trained models has fundamentally transformed the field of artificial intelligence, unlocking new possibilities and achieving unprecedented performance across a wide range of tasks.
Yet, these models inherit a fundamental limitation from traditional Machine Learning approaches: their heavy reliance on the \textit{i.i.d.} assumption, which restricts their capacity to adapt to dynamic, real-world scenarios.
We argue that the next major breakthrough in AI lies in developing systems capable of efficient and continual adaptation to evolving environments, where new data and tasks arrive sequentially.
This need defines the domain of Continual Learning (CL), a Machine Learning paradigm aimed at developing neural models capable of learning throughout their lifespan without forgetting previous knowledge.
In parallel, Parameter-Efficient Fine-Tuning (PEFT) methods have emerged as powerful tools for adapting large models to specific tasks with minimal computational resources.
While PEFT techniques can match the performance of full fine-tuning with significantly fewer parameter updates, they remain vulnerable to \textit{catastrophic forgetting}.
This survey bridges the gap between CL and PEFT, focusing on the emerging field of Parameter-Efficient Continual Fine-Tuning (PECFT).
We first provide a comprehensive overview of CL strategies and PEFT approaches, then review recent advances in PECFT.
We analyze existing methods, compare evaluation protocols and identify open challenges and promising directions for future research.
Our aim is to underscore the synergy between CL and PEFT, offering insights and guidance to researchers seeking to build adaptive, scalable and efficient AI systems.

\end{abstract}


\end{frontmatter}

\section{Introduction}
Deep learning models achieve impressive results by scaling their parameters and the data used during training. However, these models often require substantial resources, which makes it impractical to update them continuously. Although \textbf{Large Pretrained Models} (PTMs) \cite{devlin2019bert,dosovitskiy2021an,liu2019roberta} demonstrate remarkable capabilities by leveraging extensive pretraining, which allows them to perform extremely well in specific domains, {their adaptability in dynamic, continuously evolving environments remains limited.}
When deployed in real-world scenarios, fine-tuning these models becomes increasingly complex and computationally prohibitive, particularly as the number of tasks grows and data distributions continue to shift over time.

Consider a medical AI system that must continuously learn about new diseases, treatment protocols, and diagnostic techniques as medical knowledge advances. Traditional approaches would require either: (1) retraining the entire model from scratch, a computationally prohibitive and expensive process, or (2) fine-tuning on new data, which often leads to catastrophic forgetting of previously learned knowledge. Neither solution is practical for real-world deployment where systems must remain both current and comprehensive.

This challenge sits at the intersection of two fundamental limitations in current AI systems. First, while large pre-trained models excel at individual tasks, they struggle with sequential learning scenarios where new tasks arrive over time. Second, the enormous computational cost of updating these models makes frequent adaptation impractical, creating a tension between model capability and deployment efficiency.

{Two major research directions, Parameter-Efficient Fine-Tuning (PEFT) and Continual Learning (CL), have been developed to cope with some aspects of these challenges, yet each has progressed independently and is insufficient on its own to meet the demands of continuous adaptation in large pretrained models.
CL focuses on enabling models to learn from sequential tasks without forgetting previous knowledge, addressing the stability-plasticity dilemma that has long challenged machine learning systems.
Yet, a large number of traditional CL methods are developed under classic experimental settings that assume full access to all model parameters, often involving the fine-tuning of the entire network. Such assumptions make these methods computationally expensive, and the empirical conclusions drawn under these settings are not directly transferable to PTMs. As a result, their applicability to continual adaptation scenarios involving PTMs is typically limited. Nevertheless, many of the underlying methods and ideas remain insightful, although further adaptation or task-specific modifications are necessary to be effectively applied to PTMs.
On the other hand, the PEFT community offers methods to adapt large pre-trained models using only a small subset of parameters, dramatically reducing computational costs while maintaining performance. 
However, it usually assumes access to all training data simultaneously and lacks mechanisms to prevent forgetting when tasks are presented sequentially.}


To this end, combining the strengths of both endeavors leads to \textbf{Parameter-Efficient Continual Fine-Tuning} (PECFT), a new paradigm that combines the sequential learning capabilities of CL with the computational efficiency of PEFT. PECFT methods enable large pre-trained models to continuously evolve and adapt to new tasks while preserving past knowledge and maintaining computational efficiency. 
On one hand, CL enables models to continually adapt whereas PEFT ensures efficiency during adaptation, and critically, their combination resolves shortcomings that neither direction can address in isolation.

Early works in this direction \cite{wang2022learning,wang2022dualprompt} 
have demonstrated the potential of this convergence. These methods show that carefully designed parameter-efficient adaptations can enable continual learning without the traditional trade-offs between forgetting and computational cost. However, the field lacks a comprehensive understanding of the various approaches, their relative strengths and limitations, and the fundamental principles that govern effective PECFT.

The concept of PECFT represents more than just a technical advancement, it addresses a critical need for sustainable AI systems that can adapt and grow throughout their deployment lifecycle. 
Traditional machine learning paradigms assume static learning scenarios with fixed datasets and tasks. In contrast, real-world AI systems must operate in dynamic environments where:
\begin{itemize}
\item \textbf{New tasks emerge continuously} (e.g., new product categories in e-commerce, emerging medical conditions);
\item \textbf{Data distributions shift over time} (e.g., changing user preferences, evolving social media trends);
\item \textbf{Computational resources are constrained} (e.g., edge devices, cost-sensitive deployments);
\item \textbf{System downtime for retraining is unacceptable} (e.g., autonomous vehicles, medical diagnosis systems).
\end{itemize}
PECFT offers a path toward AI systems that can meet these real-world demands by enabling efficient, incremental adaptation without compromising on performance or forgetting previous capabilities.

{The growing interest in combining Continual Learning with large pre-trained models has led to the publication of several related review papers. Specifically, there are comprehensive surveys on general CL methods, reviews focused solely on Parameter-Efficient Fine-Tuning techniques for single-task adaptation \cite{han2024parameterefficientfinetuninglargemodels}, and recent, important surveys dedicated to Continual Learning with Pre-Trained Models (CL-PTM) \cite{zhou2024continuallearningpretrainedmodels} and Continual Learning for Large Language Models (CL-LLM) \cite{wu2024continuallearninglargelanguage}. Furthermore, the challenges of continual adaptation in specific domains like Vision-Language Models (VLMs) are analyzed in dedicated works \cite{liu2025continuallearningvlmssurvey}. While these CL-PTM, CL-LLM, and VLM-CL surveys acknowledge the use of PEFT methods (such as Prompts or Adapters) as part of their broader taxonomy, they do not center their analysis on the \textit{parameter efficiency} challenge itself. Our work fills this critical void. This survey is the first to comprehensively define and survey the Parameter-Efficient Continual Fine-Tuning paradigm. We differentiate ourselves by providing a structured taxonomy and analysis focused entirely on the unique trade-offs that arise when minimizing trainable parameters while maximizing knowledge retention.
}

In this survey, we provide a comprehensive overview of the PECFT research area, including motivations behind CL and PEFT, how PECFT addresses limitations in traditional approaches, existing PECFT methods, a comparison of these methods, and future directions for PECFT development.
To our knowledge, this is the first systematic and structured overview of this emerging field.
With this work, we aim to:
 \begin{itemize}

\item Foster a deeper understanding of PECFT concepts, its significance, and its role in advancements within Natural Language Processing (NLP) and Computer Vision (CV).
\item Provide a clear picture of the current state-of-the-art methodologies and applications employed in PECFT.
\item Spark exploration and discovery of promising future research directions within the field.
 \end{itemize}

The organization of this paper is as follows: 
we introduce background concepts in Sec. \ref{Background}, present a state-of-the-art and elaborated taxonomy of representative CL methods in Sec. \ref{Methodological Approach}, show an overview of current PEFT approaches in Sec. \ref{PEFT}, describe the confluence of CL and PEFT in Sec. \ref{The Confluence of CL and PEFT}, and discuss future directions in Sec. \ref{Future Directions}. Finally, we conclude this
paper in Sec. \ref{Conclusion} by summarizing our findings.

\section{Background}\label{Background}
\subsection{Continual Learning}
Continual Learning, also known as incremental or lifelong learning \cite{chen2018lifelong, aljundi2017expert, chaudhry2019efficientlifelonglearningagem, parisi2019continual, rannen2017encoder, YANG2025108226}, is a subfield of machine learning concerned with incrementally acquiring, updating, and retaining knowledge over a sequence of tasks presented over time \textcolor{red}{\cite{iparisi2019continual}}. Unlike traditional machine learning, which works with static data distributions, continual learning tackles the challenge of learning in a dynamic and evolving environment where new data often come from different distributions. 

More formally, CL considers a stream of \( T \) tasks. Each task \( t \) comprises a new dataset \( D^t = (X^t,Y^t) \), where \( X^t \) denotes the input instances and \( Y^t \) denotes the instance labels. The objective is to train a model \( f_{\Theta}: X \longrightarrow Y \) using data from a sequence of \( T \) tasks: \( D = \{D^1, ..., D^T \} \), and where each $D^t$ can follow a different distribution. Here, $\theta$ are the learned parameters of the model. During each task, the model \( f_{\Theta} \) minimizes the objective function \( \mathcal{L} \) using data \( D^t \). Each task is presented sequentially to the model and trained for $E$ epochs.
The objective function is defined as follows:

\begin{equation}
    \mathcal{L}(D^t) = \frac{1}{N^t} \sum_{t=1}^{N^t} \mathcal{L}_t (f_{\Theta}(x_i^t), y_i^t)
    \label{eq:cl}
\end{equation}

\subsubsection{CL Challenges}
Traditionally, machine learning models are usually trained on a finite, limited and static data set, limiting their adaptability. Naively training a model on new data to perform well on novel tasks can degrade its performance on previously learned tasks.
This problem is known as catastrophic forgetting \cite{mccloskey1989catastrophic} and is the core challenge of CL. Theoretically, consequential re-training and adapting to new tasks can shift how features are represented within the model. This phenomenon, where the representation of features within a model evolves in a way that can negatively impact previous tasks, is known as representation drift \cite{caccia2022new, hurtado2023continual, YANG2025108226}.

In CL, a model needs to learn corresponding task(s) with no or limited access to old training samples and perform well on both new and previous sets. Due to the need to adapt to the new task, CL approaches must find a balance between \textbf{plasticity}, the model's ability to adapt to new information and \textbf{stability}, its capacity to retain past knowledge  \cite{kim2023stabilityplasticitydilemmaclassincrementallearning}.

\subsubsection{CL Scenarios}
CL scenarios provide contexts for how the model should learn and adapt to a continuous stream of data over time; these scenarios are often categorized based on how information is presented and how the model is expected to perform.
Table \ref{tab:cl_scenarios} presents an overview of such scenarios, with their main properties.

\begin{table}[ht]
\small
\caption{Comparison of Continual Learning Scenarios}
\label{tab:cl_scenarios}
\begin{tabular}{p{29mm}|p{37mm}|p{42mm}}
\hline
\textbf{Scenario} & \textbf{Description} & \textbf{Key Features} \\
\hline
Task Incremental Learning (TIL) & Unique data distributions and non-overlapping labels per task & • Task IDs available in testing\newline • Non-overlapping class sets between tasks \\
\hline
Class Incremental Learning (CIL) & Similar to TIL, but without task information & • No task IDs in testing\newline • Must classify across all previously seen classes \\
\hline
Domain Incremental Learning (DIL) & Tasks share the same classes across different domains & • Shared label space\newline • Different data distributions\newline • No task IDs in testing \\
\hline
Online Continual Learning (OCL) & Real-time learning from continuous data stream & • Single-pass learning\newline • Immediate data disposal\newline • Streaming setup \\
\hline
Class-Incremental with Repetition (CIR) & CIL variation. Allows both new classes and repetition of previous ones & • Natural class recurrence\newline • Instance repetition\newline • Varying frequencies\\ 
\hline
Rainbow Memory (RM) & CIL variation. Addresses ``blurry'' task boundaries with shared classes & • Diversity-aware memory\newline • Enhanced augmentation\newline • Uncertainty-based selection \\
\hline
\end{tabular}
\end{table}

Some scenarios provide task identifiers during testing, allowing the model to know which task it is performing 
. One of such scenarios is \textbf{Task Incremental Learning} (TIL). In TIL, the model is trained on a sequence of tasks $T$ where each has a unique dataset and non-overlapping labels, represented as \( D^t = (X^t, Y^t, t) \) for \( t \) in \( T \), where \( p(X_i) \neq p(X_j) \) and \( Y_i \cap Y_j = \emptyset \) for \( i \neq j \). Task-specific information $t$ is available during training and testing, represented as \( p(X^t) \) for \( t \) in \( T \)  \cite{van2019three}.

A second scenario is \textbf{Class Incremental Learning} (CIL) which, while similar to TIL in training, is more challenging in inference. This is because, during testing, the specific task information is unavailable, causing the model to have to predict among all observed classes, not only those of the task $t$. This lack of task information during testing requires the model to differentiate and classify all previously learned classes without additional context, making CIL significantly more challenging. Some variations of CIL include :

\begin{enumerate}
    \item \textbf{Class-Incremental Learning with Repetition} (CIR) \cite{hemati2023class} represents a more flexible and realistic scenario in CL, where both the introduction of new classes and the repetition of previously seen classes are allowed. In CIR, the model $f_\theta$, with $\theta$ representing its parameters, learns from a stream of $N$ experiences $S = \{e_1, e_2, ..., e_N\}$, where each experience $e_i$ brings a dataset of examples $D_{e_i} = \{X_i, Y_i\}$. Unlike CI scenarios, where $Y_i \cap Y_j = \emptyset$ for $i \neq j$, or DI scenarios, where $Y_1 = ... = Y_N = Y$, CIR allows $|Y_i \cap Y_j| \geq 0$. This flexibility enables both instance repetition ($|X_i \cap X_j| \geq 0$) and concept repetition. Importantly, in CIR, repetition is a property of the environment and cannot be controlled by the CL agent, which distinguishes it from structured Replay strategies. CIR streams can be generated using methods like the Slot-Based Generator ($G_{slot}$) or the Sampling-Based Generator ($G_{samp}$), which allow for the creation of customized streams with varying degrees of repetition.

Unlike CI or DI scenarios, CIR better mimics real-world data streams where concepts naturally reoccur over time with varying frequencies. This property makes CIR particularly important for several reasons: (i) it challenges CL algorithms to balance stability and plasticity more dynamically, as they must retain knowledge of infrequent classes while adapting to frequent ones. (ii) CIR allows for the study of knowledge accumulation and refinement over time, which is critical for long-term learning systems. Potential applications of CIR are numerous and diverse. In computer vision, CIR could be applied to object recognition systems in dynamic environments, such as autonomous vehicles or surveillance systems, where particular objects may appear more frequently than others, but all must be recognized accurately.

\item \textbf{Rainbow Memory} (RM) \cite{bang2021rainbow} is another CL scenario that addresses the challenges of more realistic, ``blurry'' task boundaries. In real-world scenarios, new tasks often share classes with previous tasks, creating a continuum rather than distinct, disjoint task boundaries. This mixed setup is more challenging and practical than the traditional disjoint CL scenario. RM works by focusing on two key strategies:
\begin{itemize}
   
    \item \textbf{Diversity-aware memory update:} RM selects samples for episodic memory based on their classification uncertainty. This uncertainty is estimated using perturbed versions of the samples:
    \begin{equation}
    u(x) = 1 - \frac{1}{T} \max_c S_c,
   \end{equation}
     where $u(x)$ is the uncertainty of sample $x$, $T$ is the number of perturbations, and $S_c$ is the number of times class $c$ is predicted as the top class across perturbations. RM then selects samples across the uncertainty spectrum, ensuring a diverse representation that includes both easily classifiable and boundary samples.
    \item \textbf{Enhanced data augmentation:} RM leverages various data augmentation techniques, including mixed-label augmentations like CutMix \cite{yun2019cutmix}, to further enhance sample diversity. This helps in creating a richer, more robust representation of past tasks.
    \end{itemize}
    The relevance of RM lies in its ability to address more realistic CL scenarios. By maintaining a diverse memory of past tasks and using augmentation to enhance this diversity, RM is better equipped to handle the gradual shift in class distributions in real-world applications. This approach significantly mitigates CF.
\end{enumerate}

\textbf{Domain Incremental Learning} (DIL) \cite{shi2023unifiedapproachdomainincremental} instead refers to the scenario where a model learns to adapt to a sequence of domains over time, each one representing a variation of the same task
.
The key challenge in DIL is that the model must perform well on new domains without forgetting what it has learned from previous ones, often without knowing which domain it is dealing with during testing. Formally, we can represent this as $D^t = (X^t, Y, t)$ for $t$ in $T$, where $p(X_i) \neq p(X_j)$ for $i \neq j$, but $Y$ remains constant across all tasks. Notably, $Y_i = Y_j$ for all $i, j$ in $T$. In DIL, task identities are not required during inference, which distinguishes it from TIL. This scenario is particularly relevant in applications where the same set of classes needs to be recognized across varying domains or conditions, such as object recognition under different lighting or environmental contexts \cite{wang2024comprehensive}.

\textbf{Online Continual Learning} (OCL)\cite{aljundi2019online}  is a CL scenario that simulates real-time learning by processing a continuous stream of data with temporally shifting distributions. In this approach, the model learns directly from the incoming data, adapting to changes over time while storing only a minimal amount of information from the stream \cite{soutif2023comprehensive}. 
OCL presents a dynamic scenario where tasks have disjoint data label spaces and training samples arrive as a continuous, one-pass data stream. We can formulate this as $D = (X_t, Y_t)$ for $t$ in $\mathbb{N}$, where $Y_i \cap Y_j = \emptyset$ for $i \neq j$. In OCL, the model must learn from each sample only once, as it becomes immediately unavailable after being processed. This constraint simulates real-world scenarios where data cannot be stored or revisited due to privacy concerns or storage limitations. The challenge in OCL lies in the model's ability to continuously and quickly adapt to new classes while maintaining performance on previously learned ones, all within the constraints of a single-pass learning paradigm. This learning scenario is particularly applicable in systems that must adapt in real-time to evolving environments or user preferences, such as online recommendation systems.

\subsection{Large Pre-trained Models}

Large Pre-trained Models are massive neural networks trained on vast amounts of data to learn general-purpose representations \cite{devlin2019bert, brown2020language}. These models, predominantly based on the Transformer architectures \cite{vaswani2017attention}, have significantly altered the machine learning landscape, particularly in natural language processing and increasingly in computer vision and multimodal applications.

The ``pre-training'' in PTMs refers to the initial training phase where the model learns to perform tasks on large corpora, such as predicting masked words or generating coherent text. This process imbues the model with a broad understanding of language structures, world knowledge, and even rudimentary reasoning capabilities \cite{petroni2019languagemodelsknowledgebases}. The ``large'' aspect denotes the model's size in terms of parameters and the scale of data and computation involved in their training \cite{kaplan2020scalinglawsneurallanguage}.

A key advantage of PTMs is their capacity to learn robust representations that serve as excellent starting points for a wide range of downstream tasks, often requiring minimal task-specific training data \cite{raffel2023exploringlimitstransferlearning}. In practice, pre-trained models are adapted to specific tasks through fine-tuning or by training new classifier layers, making them versatile tools across various domains.

Key examples of large PTMs showcase their diverse capabilities:
\begin{itemize}
    \item BERT \cite{devlin2019bert} revolutionized natural language understanding tasks.
    \item GPT-3 \cite{brown2020language} demonstrated remarkable text generation abilities.
    \item T5 \cite{raffel2023exploringlimitstransferlearning} unified various NLP tasks under a single model.
    \item Vision Transformers \cite{dosovitskiy2021an} applied similar principles to image recognition.
\end{itemize}

These models have pushed the boundaries of artificial intelligence, exhibiting capabilities that range from human-like text generation to solving complex reasoning tasks, often with little to no task-specific training \cite{wei2022emergent}. Their effectiveness in capturing knowledge from large volumes of labeled and unlabeled data, combined with their adaptability, has made PTMs crucial in multiple research and application areas.

The fundamental architecture commonly used with large PTMs is the Transformer \cite{vaswani2017attention}, built around the self-attention mechanism. Specifically, given an input sequence \( X = \{x_1, x_2, \ldots, x_n\} \), where \( x_i \) represents the \(i\)-th token or patch, the Transformer module computes attention scores using queries \( Q \), keys \( K \), and values \( V \), which are linear projections of \( X \). Self-attention mechanism is defined as:

\begin{equation}
\text{Attention}(Q, K, V) = \text{softmax}\left(\frac{QK^T}{\sqrt{d_k}}\right)V
\end{equation} 
where \( d_k \) is the dimensionality of the keys. This attention allows the model to weigh the importance of each token/patch relative to others, capturing contextual relationships efficiently.
This architectural component makes language models excel in natural language understanding and generation by employing multiple layers of self-attention and feed-forward networks. These models are trained to generate text by predicting the next token or finding the masked value in a sequence, maximizing the probability \( P(x_{i+1} | x_1, x_2, \ldots, x_i) \).

There also exist vision models, \textit{e.g.} ViT \cite{dosovitskiy2021an}, that apply the Transformer architecture to image data. An image is divided into patches, flattened and linearly embedded into a sequence of vectors. Given an image \( I \) divided into \( N \) patches, each patch \( p_i \) is embedded into a vector \( e_i \). The sequence of embedded patches \( E = \{e_1, e_2, \ldots, e_N\} \) is then processed through the transformer architecture similarly to text tokens, allowing the model to capture spatial relationships within the image.

Although PTMs have demonstrated remarkable success across various tasks, they have some limitations. One significant challenge is the domain gap between pre-training and target task data. Pre-trained models are typically trained on large-scale datasets; however, in many cases, these datasets will not fully represent the specific characteristics of the target task. This domain gap can lead to performance degradation when deploying pre-trained models directly to new tasks or domains \cite{guo2022domainadaptationgeneralizationpretrained}.

One way of alleviating this problem is to employ \textbf{Transfer Learning} strategies \cite{zhuang2020comprehensive}. Transfer learning involves leveraging knowledge from a source domain to improve performance on a target domain or task. By fine-tuning the model parameters on a smaller dataset related to the target task, the model can adapt to the specific characteristics of the task while using the pertaining as a good starting point. 

Parameter-Efficient Fine-Tuning has emerged as a significant advancement in large PTMs studies. It refines the capabilities of pre-trained models by strategically adjusting a limited subset of parameters during fine-tuning. This approach differs significantly from comprehensive fine-tuning, which requires the entire model to be retrained, often at considerable computational cost. PEFT achieves impressive results while minimizing parameter updates, ensuring efficient adaptation and preserving the model's previously acquired knowledge~\cite{fu2023effectiveness}. 

\section{Overview of Classical Continual Learning Algorithms}
\label{Methodological Approach}

\begin{figure}
    \centering
    \includegraphics[width=1\textwidth]{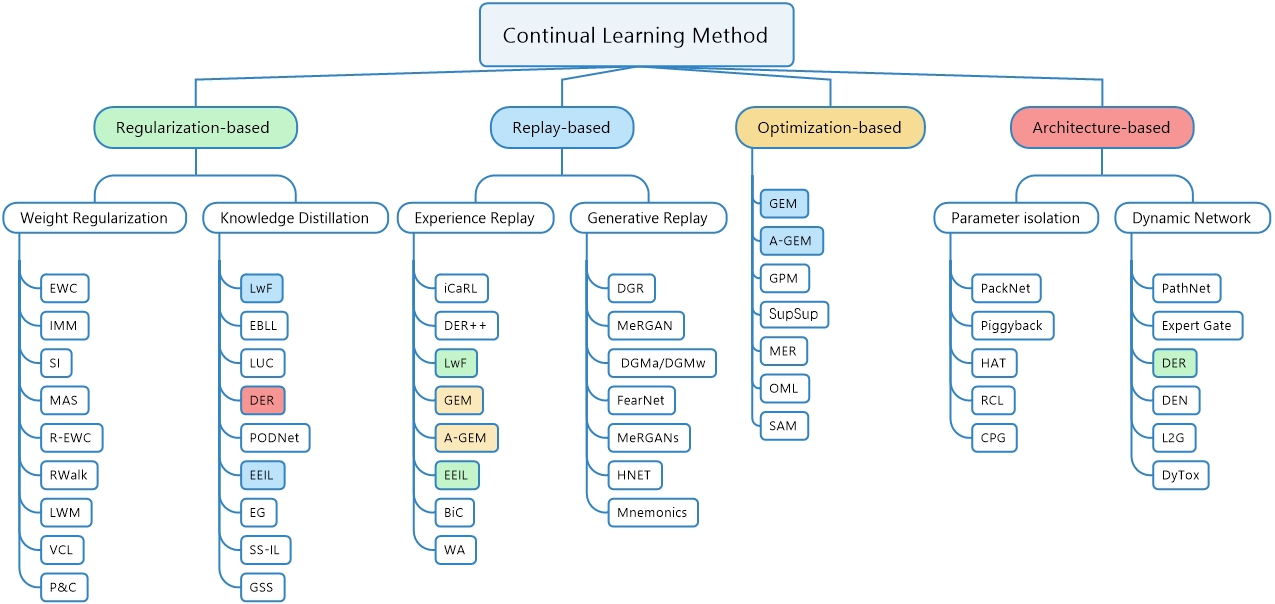}
    \caption{Overview of CL Methods. The four main categories of CL methods are represented in green, blue, yellow, and red, respectively. Note that some methods may belong to multiple categories.}
    \label{fig:CL Method}
\end{figure}

Continual Learning is learning from dynamic data distributions arriving in sequence. 
As shown in Figure \ref{fig:CL Method}, CL is generally divided into four categories: regularization-based, replay-based, optimization-based, and architecture-based approaches; for more detail, see previous work that explains the differences in these types of approaches \cite{wang2024comprehensive, zhou2024class, ijcai2024p924, YANG2025108226}.
{The following sections offer an overview of each category, emphasizing key features and representative methods.}

\subsection{Regularization-based CL} 
An alternative to naively training a model in a sequence of tasks is adding constraints to the loss functions to encourage mitigation of forgetting. This regularization can be imposed on weights by estimating the importance of the parameters so that the relevant weights do not drift significantly. One of the first approaches that proposed this was \textbf{Elastic Weight Consolidation} (EWC)~\cite{Kirkpatrick_2017}, which captures the prior importance using a diagonally approximated Fisher information matrix.

Regularization can also be imposed on activations to prevent activation drift, which generally outperforms its weight-regularisation counterpart. One work in this line is \textbf{Learning without Forgetting} (LwF)~\cite{li2017learning}, which prevents activations of the old network from drifting while learning new tasks. The less-forgetting learning penalizes the activation difference except for the fully-connected layer. \textbf{Riemannian Walk} \cite{chaudhry2018riemannian} extends EWC by incorporating KL-divergence-based regularization, path integral methods to assess parameter importance and sample-based strategies for retaining past knowledge. \textbf{Rotated EWC} (R-EWC) \cite{liu2018rotate} enhances EWC by reparameterizing the network to better align with the Fisher information matrix, improving the diagonal approximation and reducing forgetting.

Similarly, \textbf{Synaptic Intelligence} (SI) \cite{zenke2017continual} evaluates the importance of each synapse online during training and consolidates significant parameters by penalizing considerable changes to mitigate forgetting. \textbf{Memory Aware Synapses} (MAS) \cite{aljundi2018memory} determine the importance of weights based on the sensitivity of the learned function, enabling adaptive penalization of changes to significant weights without relying solely on the loss function. Efforts have also been directed towards enhancing the implementation of secondary penalties. \textbf {Incremental Moment Matching} (IMM) \cite{lee2017overcoming} integrates multiple models trained on different tasks by aligning their weight distributions, employing weight transfer, and L2 regularization to sustain performance across tasks.

Function regularization is another way, also known as \textbf{Knowledge Distillation} (KD) \cite{hinton2015distillingknowledgeneuralnetwork}, to target the intermediate or final output of the prediction function. This approach typically employs the previously-learned model as the teacher and the currently-trained model as the student, leveraging knowledge distillation techniques to mitigate catastrophic forgetting. In ideal scenarios, KD would target all old training samples, but in continual learning settings, alternatives such as new training samples, a small fraction of old training samples, external unlabeled data, or generated data are used, albeit suffering from varying degrees of distribution shift. 

Pioneering works like \textbf{iCaRL} \cite{rebuffi2017icarl} learn new training samples while utilizing predictions from the output heads of the old tasks to compute the distillation loss. \textbf{LwM} \cite{dhar2019learningmemorizing} exploits the attention maps of new training samples for KD, while \textbf{EBLL} \cite{rannen2017encoder} learns task-specific autoencoders to prevent changes in feature reconstruction. \textbf{GD} \cite{lee2019overcoming} further distills knowledge from the large stream of unlabeled data available. 

{Zhou et. al.~\cite{zhou2024class} further categorize knowledge distillation--based methods into three subgroups: \textbf{logit distillation}, \textbf{feature distillation}, and \textbf{relational distillation}. Logit distillation aims to transfer knowledge by encouraging consistency between the output logits of the previous and current models. A seminal example of this approach in the continual learning literature is \textbf{Learning without Forgetting (LwF)}~\cite{li2017learning}, which leverages the predictions of the old model to regularize the training of the new model and mitigate catastrophic forgetting.}

Feature distillation ensures the protection of learned knowledge by distilling at the intermediate levels of the models. Methods like \textbf{UCIR} \cite{hou2019learning} force the features extracted by the new embedding module to be the same as the old one, providing a stronger regularization. Other works like \textbf{LwM} \cite{dhar2019learning} and \textbf{AFC} \cite{Kang2022afc} utilize attention maps and feature importance, respectively.

Relational distillation introduces a different perspective by distilling the relationships between multiple samples or instances. This level captures the structural information within the data, enhancing the retention of old knowledge. Methods like \textbf{COIL} \cite{zhou2021co} suggest bidirectional distillation with co-transport, utilizing semantic relationships between old and new models.

\subsection{Replay-based CL} 
Replay-Based methods store a small subset of data from the previously accessed tasks to reinforce the network’s memory of old knowledge. Current memory-based methods have achieved promising results on many CL benchmarks. Saving examples helps mitigate forgetting of previous tasks by representing past distributions used during training. For the memory to be sufficient, it must represent the previous distribution as fully as possible, considering all its classes and concepts~\cite{YANG2025108226}.

For the case of replay-based methods, together with minimizing Equation \ref{eq:cl}, model $f_{\Theta}$ needs to minimize $\mathcal{L}$ using the data available in memory $M$ at time $t$. The buffer $M^t$ comprises $|M|$ samples from previous distributions, meaning that at task $t$, the buffer will contain samples only from $t'<t$, as shown in Equation \ref{eq:cl_memory}:

\begin{equation}
    \mathcal{L}_M(D^t, M^t) = \mathcal{L}(D^t) + \frac{1}{|M|} \sum_{i=1}^{|M|} \mathcal{L}_t (f_{\Theta}(m_i^t, y_i^t) 
    \label{eq:cl_memory}
\end{equation}
The function in charge of populating memory $M$ is known as \textbf{Storage Policy} \cite{hurtado2023memory} and decides which elements go into the memory by sampling from set $D^t$ given a function $\mathbf{P}$, as shown in Equation \ref{eq:storage_policy}. An ideal policy function is the one that minimizes Equation \ref{eq:cl_memory} for evaluation stream $D^1 ... D^T$, restricted by the memory size $|M|$. 

\begin{equation}
    M^{t+1} \leftarrow \mathbf{P}(M^t, D^t), \quad |M^{t+1}| \leq |M|
    \label{eq:storage_policy}
\end{equation}

In most cases, we assume that $M^{t+1}$ will always contain $|M|$ samples, and the storage policy will decide which samples to remove to add those from new task $D^t$.
In practice, designing an effective storage policy involves striking a balance between preserving diversity in the memory, \textit{i.e.} representing past distributions fully, and accommodating new information from the current task. This delicate balance ensures that the memory remains relevant and informative over time, contributing to the model's overall performance across the entire sequence of tasks. 
The replay-based CIL branch quickly draws attention due to its appealing ability to resist catastrophic forgetting. For instance, \textbf{PODNet} \cite{douillard2020podnet} adopts an efficient spatial-based distillation loss to reduce forgetting, with a focus on the large-phase setting, achieving reasonably good results.
\textbf{AANets} \cite{Liu_2021} employs a new architecture containing a stable block and a plastic block to balance the stability and plasticity.\\
On top of the replay-based CIL, methods exploring exemplar storing techniques are also fruitful. For instance, \textbf{Reinforced Memory Management} (RMM) \cite{liu2023rmm} seeks dynamic memory management using reinforcement learning. By plugging it into PODNet and AANets, RMM attains a state-of-the-art performance.\\
GAN-based CIL replays past samples by generating them using GANs. The deep generative replay generates synthetic samples using an unconditioned GAN. It is later improved by memory replay GAN adopting a label-conditional GAN. GAN-based CIL relies heavily on GAN’s generative performance and is only tested on relatively small datasets, such as MNIST.
Bias correction-based CIL mainly tries to address the task-recency bias. The end-to-end incremental learning reduces the bias by introducing a balance training stage where only an equal number of samples for each class is used. The \textbf{Bias Correction} (BiC) \cite{wu2019large} includes an additional trainable layer which aims to correct the bias. The method named \textbf{LUCIR} \cite{Hou_2019_CVPR} fights the bias by changing the softmax layer into a cosine normalization one.

\subsection{Architecture-Based CL}
In architecture-based methods of CL, the network architecture is dynamically updated during the learning process to retain previously acquired knowledge. These methods are designed to adjust the model’s architecture to effectively handle new tasks while maintaining the knowledge from earlier tasks. Architecture-based methods can be categorized into fixed-capacity and capacity-increasing approaches based on the evolution of their model parameters as the number of tasks increases~\cite{YANG2025108226}. Building on these foundations, we explore the implementation of task-specific parameters, extending these concepts through parameter allocation and dynamic network architectures.

Parameter allocation approaches dedicate specific and isolated parameter subspaces within the network to each task. \textbf{Piggyback} \cite{mallya2018piggyback}, \textbf{HAT} \cite{duan2024hatclhardattentiontothetaskpytorchlibrary}, \textbf{WSN} \cite{pmlr-v162-kang22b}, and \textbf{H2} \cite{10.1007/978-3-031-20083-0_31} utilize binary masks to select specific neurons or parameters for each task, effectively freezing the old tasks' masked regions to prevent forgetting. \textbf{PackNet} \cite{mallya2018packnetaddingmultipletasks}, \textbf{CLNP} \cite{golkar2019continuallearningneuralpruning}, and \textbf{AGS-CL} \cite{jung2021continuallearningnodeimportancebased} identify important neurons or parameters for the current task and release the rest for subsequent tasks. However, limited network capacity can lead to saturation of ``free'' parameters as the number of tasks increases. Dynamic architecture expansion can address this by allowing the network to grow when necessary, with methods like reinforcement learning, architecture search, and variational Bayes being employed for optimization.

Dynamic network architectures can be further divided into model decomposition approaches and modular network.\\
{Model Decomposition: }Methods employing model decomposition explicitly separate a model into task-sharing and task-specific components. Regular networks can incorporate task-specific components as parallel branches, adaptive layers, or masks of intermediate features. Additionally, network parameters themselves can be decomposed into task-sharing and task-specific elements. While this approach offers scalability, the number of task-specific components typically grows linearly with tasks, making resource efficiency a crucial factor.\\
{Modular Networks: }Modular networks utilize parallel sub-networks or sub-modules to learn tasks independently. Early works like \textbf{Progressive Networks} \cite{rusu2016progressive} introduced an identical sub-network for each task, facilitating knowledge transfer through adapter connections. \textbf{Expert Gate} \cite{aljundi2017expert} employs a mixture of experts, expanding one for each new task. \textbf{PathNet} \cite{fernando2017pathnet} pre-allocates parallel networks to construct candidate paths, selecting the best path for each task. \textbf{MNTDP} \cite{veniat2021efficientcontinuallearningmodular} seeks to find the optimal layout from existing and potentially new sub-modules. Similar to Parameter Allocation, these methods aim to construct task-specific models while enabling explicit knowledge reuse through sub-network combinations. 

Unlike other directions, most architecture-based approaches have the objective to de-correlate tasks in network parameters, potentially sacrificing scalability and inter-task generalizability~\cite{YANG2025108226}. Task identities are often required to determine which set of parameters to use. This limitation can be mitigated by inferring task identities from the predictive uncertainty of task-specific models or by learning the function of task-identity prediction through continual learning strategies.

\subsection{Optimization-based CL} 
Beyond incorporating additional loss terms, \textit{e.g.} regularization and replay, the optimization-based approach in CL explores alternative optimization strategies. {Previous surveys \cite{wang2024comprehensive, YANG2025108226} have systematically examined the boundaries, distinctions, and interactions between regularization-based and optimization-based methods, highlighting how each contributes to mitigating catastrophic forgetting from different perspectives.} Building on these insights, recent studies have introduced techniques such as \textbf{gradient projection} to constrain model updates and prevent knowledge loss \cite{Zeng_2019, farajtabar2019orthogonalgradientdescentcontinual,wang2023orthogonalsubspacelearninglanguage, saha2021gradientprojectionmemorycontinual,lin2022trgptrustregiongradient}, as well as \textbf{meta-learning} approaches that aim to automatically acquire inductive biases suitable for CL scenarios \cite{javed2019metalearningrepresentationscontinuallearning, beaulieu2020learningcontinuallylearn, riemer2019learninglearnforgettingmaximizing, rajasegaran2020itamlincrementaltaskagnosticmetalearning, gupta2020lamamllookaheadmetalearning}.

CL encompasses a diverse range of techniques that go beyond regularization and experience replay to address the challenge of learning from a continuous stream of data without forgetting previously acquired knowledge.
Gradient projection methods, such as \textbf{GEM} \cite{lopezpaz2022gradientepisodicmemorycontinual}, \textbf{A-GEM }\cite{chaudhry2019efficientlifelonglearningagem}, and \textbf{LOGD }\cite{tang2021layerwiseoptimizationgradientdecomposition}, manipulate parameter updates to align with or remain orthogonal to specific directions, preserving past input and gradient spaces.
In contrast, \textbf{OWM} \cite{Zeng_2019} and \textbf{OGD} \cite{farajtabar2019orthogonalgradientdescentcontinual} offer contrasting strategies: OWM modifies updates to be orthogonal to the previous input space, while OGD preserves old gradient directions and rectifies current ones. \textbf{OrthogSubspace} \cite{wang2023orthogonalsubspacelearninglanguage} and \textbf{GPM} \cite{saha2021gradientprojectionmemorycontinual} leverage orthogonal subspaces for CL, with \cite{saha2021gradientprojectionmemorycontinual} focusing on maintaining the gradient subspace of old tasks.
\textbf{FS-DGPM} \cite{deng2021flatteningsharpnessdynamicgradient} dynamically adjusts \cite{saha2021gradientprojectionmemorycontinual} by discarding unimportant bases to enhance learning plasticity and convergence. \textbf{TRGP} \cite{lin2022trgptrustregiongradient} defines a ``trust region'' based on gradient projection to selectively reuse frozen weights from prior tasks.

Meta-learning approaches in CL aim to acquire a data-driven inductive bias suitable for various scenarios, eliminating the need for manual design. \textbf{OML} \cite{javed2019metalearningrepresentationscontinuallearning} employs meta-training to perform online updates while minimizing interference, promoting sparse representations. \textbf{ANML} \cite{beaulieu2020learningcontinuallylearn} extends this idea by meta-learning a context-dependent gating function to activate relevant neurons for each task.
Meta-learning can be combined with experience replay for better utilization of both old and new data. \textbf{MER} \cite{riemer2019learninglearnforgettingmaximizing} aligns gradient directions, while \textbf{iTAML} \cite{rajasegaran2020itamlincrementaltaskagnosticmetalearning} applies a meta-updating rule for balancing them. \textbf{La-MAML} \cite{gupta2020lamamllookaheadmetalearning} optimizes the OML objective with an adaptive learning rate and leverages replay for online training. \textbf{OSAKA} \cite{caccia2021onlinefastadaptationknowledge} proposes a hybrid objective for knowledge accumulation and fast adaptation, achieved through meta-training initialization and task incorporation. \textbf{MERLIN} \cite{joseph2020metaconsolidationcontinuallearning} utilizes a metadistribution of model parameters to sample and ensemble task-specific models for inference, while \textbf{MARK} \cite{hurtado2021optimizingreusableknowledgecontinual} maintains incrementally updated shared weights through meta-learning and selective masking for specific tasks. 

\begin{table}[htbp]
\centering
\caption{Overview of Parameter-Efficient Fine-tuning Methods}
\label{tab:peft_methods}
\begin{adjustbox}{max totalheight=0.96\textheight}
\begin{tabular}{p{38mm} p{74mm}}
\toprule
\textbf{Method} & \textbf{Key Mechanism and Features} \\
\midrule
\multicolumn{2}{l}{\textbf{Adapter-Based Methods}} \\
\midrule
Adapter \cite{pmlr-v97-houlsby19a} & Bottleneck layers (2kd/layer) with down/up projections. Effective task adaptation with minimal overhead \\
AdaptFormer \cite{chen2022adaptformer} & Specialized vision modules enhancing ViT transferability across tasks \\
Adapter Fusion \cite{pfeiffer2021adapterfusion} & Two-stage fusion combining task adapters with knowledge distillation \\
Mera \cite{he2023mera} & Efficient integration of existing adapters using same-track combination strategy \\
AdaMix \cite{wang2022adamix} & Multiple parallel adapters providing diverse task perspectives \\
Invertible Adapters \cite{pfeiffer-etal-2020-mad} & Reversible bottleneck modules using NICE-style coupling for zero-shot cross-lingual transfer \\
Adapter+ \cite{steitz2024adaptersstrike} & Lightweight variant with channel-wise scaling and optimized placement (only after multi-head attention) \\
Parallel Adapters \cite{he2022towards} & Run alongside feedforward sublayers rather than sequentially, enabling parallel computation \\
Mix-and-Match Adapters \cite{he2022towards} & Combine prefix tuning with scaled parallel adapters\\
\midrule
\multicolumn{2}{l}{\textbf{Prompt-Based Methods}} \\
\midrule
Prompt Tuning \cite{lester2021power} & Continuous trainable prefix tokens (l×d) for task conditioning \\
P-Tuning \cite{liu2022p} & Hybrid discrete and continuous prompts with LSTM/MLP encoder \\
Prefix Tuning \cite{li2021prefix} & Optimizable virtual tokens (0.1\%) as prefix for generation \\
MPT \cite{wang2023multitaskprompttuningenables} & Distillation-based transferable prompts (0.035\%) for multi-task scenarios \\
\midrule
\multicolumn{2}{l}{\textbf{Reparameterization Methods}} \\
\midrule
LoRA \cite{hu2022lora} & Low-rank decomposition (2dr/layer) for weight updates. Zero inference overhead \\
QLoRA \cite{dettmers2023qlora} & 4-bit NormalFloat quantization + LoRA (0.1\%) for memory-efficient training \\
DoRA \cite{liu2024dora} & Separates magnitude and direction components for improved training dynamics \\
VB-LoRA \cite{li2024vb} & Constructs low-rank matrices from a shared vector bank \\
VeRA \cite{kopiczko2024veravectorbasedrandommatrix} & Shared frozen matrices with learnable scaling vectors \\
\midrule
\multicolumn{2}{l}{\textbf{Selective Methods}} \\
\midrule
BitFit \cite{zaken2021bitfit} & Selective bias-term updates ($< 0.1\%$) preserving model knowledge \\
Masking \cite{zhao2020masking} & Task-specific binary masks identifying crucial weights (3-10\%) \\
\bottomrule
\end{tabular}
\end{adjustbox}
\end{table}

\section{Parameter-Efficient Fine-Tuning} \label{PEFT}
Fine-tuning is a transfer learning technique that adapts a pre-trained model to a specific downstream task by further training it on task-specific data \cite{devlin2019bert}. This process typically involves updating all or most of the model's parameters to optimize performance on the new task. Fine-tuning has become crucial in modern ML pipelines, particularly for PTMs, as it allows leveraging the rich representations learned from vast amounts of data to solve specific tasks with relatively small datasets \cite{raffel2023exploringlimitstransferlearning}. The importance of fine-tuning lies in its ability to significantly reduce training time and computational resources compared to training models from scratch while often achieving superior performance \cite{howard2018universal}.

However, as PTMs grow in size, with some models containing billions of parameters, traditional fine-tuning becomes increasingly challenging and expensive. This approach often requires specialized hardware, substantial energy consumption \cite{kaplan2020scalinglawsneurallanguage, brown2020language} and can lead to potential overfitting on smaller downstream datasets \cite{strubell2019energy}. Moreover, full fine-tuning can result in catastrophic forgetting, where the model loses its ability to perform well on previously learned tasks \cite{kirkpatrick2017overcoming}.

In the context of large-scale PTMs, Parameter-Efficient Fine-Tuning  has emerged as a resource-efficient approach for model adaptation \cite{he2022towards}. PEFT methods aim to achieve performance comparable to or even surpass full model fine-tuning while updating only a small number of trainable parameters, either by selectively updating a subset of the model's parameters \cite{zhao2020masking} or introducing new task-specific parameters \cite{pmlr-v97-houlsby19a}. This approach significantly reduces computational costs and memory requirements, making it particularly effective when working with very large models or in scenarios with limited computational resources, as it enables efficient adaptation across various tasks without extensive retraining.

PEFT strategies can be broadly categorized based on how they modify or add parameters to the pre-trained model. These categories include: (i) additive methods, which introduce new task-specific parameters to the model; (ii) reparameterization methods, which reparameterize the existing model parameters more efficiently; (iii) and selective methods, which fine-tune a subset of the model parameters based on their importance to the task. Since this paper focuses on the intersection of Continual Learning and Parameter-Efficient Fine-tuning, we will present an overview of these methods, introducing and discussing only a few of these , for a comprehensive comparison of different PEFT categories and methods, we refer readers to recent surveys by \cite{ijcai2024p924} and \cite{xin2024parameterefficient}.
A summary of the PEFT methods we discuss can be seen in Table \ref{tab:peft_methods}.
Following the overview of PEFT techniques, we analyze the computational efficiency of each approach, focusing on the number of trainable parameters introduces across various backbone models. To ensure a broad and comprehensive comparison, we include models from both NLP and Computer Vision domains, as well as a multi-modal backbone.
This comparison, presented in Table \ref{tab:peft_comparison}, offers valuable insights for practitioners, providing efficiency metrics aside method descriptions to facilitate a clearer understanding of each PEFT strategy.

\subsection{Additive Methods}
In additive methods, task-specific parameters and lightweight modules are added to the pre-trained model architecture.

\subsubsection{Adapter Based PEFT Methods}
In the original \textbf{Adapter} method \cite{pmlr-v97-houlsby19a}, small neural network modules , the so-called ``adapters'' , are inserted into the transformer layers of the pre-trained model. These adapters are then trained on the downstream task while keeping the pre-trained model parameters frozen. Adapters have a bottleneck structure, and each adapter block comprises two fully connected layers: the first layer projects the input $x \in \mathbb{R}^d$ into a lower-dimensional space $z \in \mathbb{R}^k$, with $k \ll d$, while the second layer restores this representation to the original dimension. The total number of parameters introduced is $2kd$, significantly less than that of a single fully connected layer. Only the adapter block's parameters are adjusted during fine-tuning, leaving the pre-trained model parameters frozen. This ensures effective task adaptation with minimal computational overhead. 

\textbf{AdaptFormer} \cite{chen2022adaptformer} enhances pre-trained Vision Transformers for various image and video tasks with significant benefits. It employs lightweight modules, adding less than 2\% extra parameters to a ViT. This approach improves ViT's transferability without altering its original pre-trained parameters, allowing it to surpass existing fully fine-tuned models in action recognition benchmarks. Such method also offers the flexibility to combine multiple adapters, usually for different downstream tasks. 

\textbf{Invertible adapters} \cite{pfeiffer-etal-2020-mad} introduce a reversible bottleneck module inside the transformer. They use a NICE-style coupling mechanism: the input embedding is split in half and transformed by projections G, making the process invertible. This means no information is lost. They train these adapters with masked language modeling on unlabeled language data. In the MAD‑X framework, they stack invertible adapters along with language and task adapters. At inference, the English adapter is swapped out with a target-language one. This method keeps the base model frozen, adds only a small number of parameters, and supports zero-shot cross-lingual transfer—even to languages unseen during pre‑training.

\textbf{Adapter+} \cite{steitz2024adaptersstrike} presents a lightweight yet robust variant that refines classic adapter designs through careful empirical analysis. Unlike standard approaches such as \cite{pmlr-v97-houlsby19a}  Adapter+ places adapters only after multi-head attention, introduces channel-wise scaling, and adopts Houlsby-style \cite{pmlr-v97-houlsby19a} initialization. These adjustments yield strong out-of-the-box performance with minimal tuning, rivaling other adapter based methods. 

\textbf{Parallel adapters} \cite{he2022towards}  differ from standard sequential adapters by running alongside the feedforward sublayer rather than after it. The authors add the adapter output directly to the base model’s output. This structure lets the adapter learn only residual corrections. It’s more flexible and can improve performance in transfer settings. It also enables parallel computation paths for faster inference.

\textbf{Mix‑and‑Match adapters} \cite{he2022towards} combine beneficial features from various adapter approaches. They use prefix tuning at attention layers with a small bottleneck and scaled parallel adapters for feedforward layers. In their unified framework, adapters from different tasks can be combined. This variant matches full fine-tuning performance on some benchmarks while fine‑tuning only about 6.7\% of parameters. Although it is a predominantly adapter-based, but they incorporate prompt-based ideas, specifically from  \cite{li2021prefix}.

In \textbf{Adapter Fusion} \cite{pfeiffer2021adapterfusion}, a two-stage learning algorithm is used to combine knowledge from multiple tasks efficiently. This approach prevents catastrophic forgetting and avoids issues related to dataset balancing by clearly separating the stages of knowledge extraction and composition. This separation enables the classifier to effectively utilize the representations learned from various tasks, enhancing overall efficiency.  

Similarly, \textbf{Mera} \cite{he2023mera} integrates multiple pre-existing adapters into a unified model via model fusion, all while markedly enhancing the performance of \cite{pfeiffer2021adapterfusion}. This is achieved through the ``\textit{same-track}'' strategy, which combines adapters from the same pretraining task sequence.

\textbf{AdaMix} \cite{wang2022adamix} presents an approach to fine-tuning that maximizes parameter efficiency by employing a mixture of adapter modules. This technique improves performance on downstream tasks while minimizing changes to PTM weights. Additionally, AdaMix is engineered to match the computational cost and trainable parameters of the underlying PEFT method. Unlike conventional PEFT methods, which typically employ a single adaptation module per Transformer layer, AdaMix incorporates multiple adaptation modules to capture diverse perspectives of the task. Its adaptable framework can seamlessly integrate into various PEFT methods, such as \cite{pmlr-v97-houlsby19a,hu2022lora}, highlighting its versatility.

\subsubsection{Prompt Based PEFT Methods}
Another group of additive methods are soft prompts, which are trainable tensors that are concatenated to the inputs. The key idea is that such prompts can learn and adapt during fine-tuning, while the original weights of the pre-trained model remain intact. Several approaches have been developed to leverage this concept effectively.

\textbf{Prompt Tuning} \cite{lester2021power} adapts frozen, pre-trained language models to perform specific downstream tasks by introducing additional learnable prompt tokens, represented as \( P = [P_1, P_2, \ldots, P_l] \).
These tokens are concatenated with the original input \( X \in \mathbb{R}^{n \times d} \) to construct the final input \( [P; X] \).
In standard prompting, additional information is added to the input, allowing the model to maximize the likelihood of generating the correct output without altering the model parameters \(\theta\). This is typically achieved by selecting prompt tokens either manually or via non-differentiable search methods. Prompt tuning, however, decouples the prompt from the model's fixed parameters. This decoupling allows the embeddings of these prompt tokens to be learned and optimized, making it easier to adapt to specific tasks.

Building on this foundation, \textbf{P-Tuning} \cite{LIU2024208} was introduced to address a critical limitation , the inherent instability issues of discrete prompts.
It introduces trainable continuous prompt embeddings that are concatenated with traditional discrete prompts. The core template structure is defined as:

\begin{equation}
  T = \{[P_{0:i}], x, [P_{(i+1):j}], y, [P_{(j+1):k}]\}
\end{equation}
\noindent
where $[P_i]$ represents continuous prompt embeddings that are optimized through backpropagation, while $x$ and $y$ represent the input and label respectively. The combination of trainable continuous embeddings with discrete prompts is done via a prompt encoder, \textit{e.g.} LSTM or MLP, to model the dependencies between the embeddings. This hybrid approach significantly reduces performance variance, maintaining a consistent behavior where traditional methods showed up to 20\% drops from single word changes in prompts.

Similarly, \textbf{Prefix Tuning} \cite{li2021prefix} prepends trainable continuous vectors to the input, allowing the transformer to attend those as ``\textit{virtual tokens}'' while keeping the main model parameters frozen. The activations $h_i$ are computed as:
\begin{equation}
  h_i = \begin{cases}
      P_\theta[i, :], & \text{if } i \in P_{\text{idx}} \\
      \text{LM}(z_i, h_{<i}), & \text{otherwise}
  \end{cases}
\end{equation}
\noindent
where $P_\theta$ is parameterized using an MLP for stability: $P_\theta[i, :] = \text{MLP}_\theta(P'_\theta[i, :])$. This approach achieves comparable performance to full fine-tuning while only requiring 0.1\% of the parameters.

Taking prompt tuning to a multi-task setting, \textbf{Multitask Prompt Tuning} (MPT) \cite{wang2023multitaskprompttuningenables} innovates by learning a single transferable prompt from multiple source tasks that can be adapted to target tasks. By operating in two stages , source training and target adaptation , MPT learns a task-shared prompt matrix through knowledge distillation. This prompt is decomposed into a task-shared matrix $P^*$ and low-rank task-specific matrices $W_k = u_k \otimes v_k^T$ for each task $k$, with the task-specific prompt parameterized as $\hat{P}_k = P^* \odot W_k$. The method combines multiple loss functions:
\begin{equation}
  \mathcal{L}_{\text{Total}} = \mathcal{L}_{\text{PLM}} + \lambda(\mathcal{L}_{\text{Logits}} + \mathcal{L}_{\text{Hidden}})
\end{equation}
\noindent
where $\mathcal{L}_{\text{PLM}}$ represents task-specific losses, $\mathcal{L}_{\text{Logits}}$ aligns probability distributions between teacher and student models via KL-divergence, and $\mathcal{L}_{\text{Hidden}}$ minimizes differences in hidden states. This approach achieves impressive performance while tuning only 0.035\% of task-specific parameters.

\subsection{Reparameterization Methods}
Reparameterization methods provide an alternative way to represent and update a neural network's parameters during training. Instead of directly modifying the original weight matrices, it expresses them through a transformation function using a smaller set of parameters. The key advantage is that we can achieve similar model capabilities while training far fewer parameters, making the process more efficient and controllable.

\textbf{LoRA} \cite{hu2022lora} addresses the challenge of adapting massive pre-trained models to specific tasks by using low-rank decomposition matrices to indirectly update the model's weights.
It introduces a low-rank update to the pre-trained model weights, expressed as:
\begin{equation}
W = W_0 + BA
\end{equation}
where $W_0 \in \mathbb{R}^{d \times k}$ represents the initial pre-trained weight matrix, while $B \in \mathbb{R}^{d \times r}$ and $A \in \mathbb{R}^{r \times k}$ are the additional parameters, with rank $r \ll \min(d,k)$.
During finetuning, $W_0$ is kept frozen while the $A$ and $B$ matrices are updated.
This formulation allows for efficient computation and storage, as the update matrices $B$ and $A$ are much smaller than the original weight matrix.
LoRA is particularly efficient both for the low parameter count and for the small GPU usage during training. Furthermore, it introduces no additional latency during inference, as the update $\Delta W = BA$ can be merged with $W_0$ post-training.
 
\textbf{QLoRA} \cite{dettmers2023qlora} is a reparameterization approach that enables training of LLMs on limited hardware while maintaining performance. It achieves this by combining quantization with \cite{hu2022lora} through several key innovations.
QLoRA processes weight matrices through a dual approach:

\begin{equation}
  Y^{\text{BF16}} = X^{\text{BF16}}\text{doubleDequant}(c_1^{\text{FP32}}, c_2^{k\text{-bit}}, W^{\text{NF4}}) + X^{\text{BF16}}L_1^{\text{BF16}}L_2^{\text{BF16}}
\end{equation}
\noindent
where the base model weights are quantized to 4-bit precision using NormalFloat (NF4), while the LoRA adapters ($L_1, L_2$) operate in 16-bit precision.
QLoRA integrates three major components that work together to enable efficient training: \textit{NF4}, an information-theoretically optimal quantization format for normally distributed weights that outperforms standard 4-bit quantization; \textit{Double Quantization}, which further compresses the model by quantizing the quantization constants themselves, reducing memory requirements by approximately 0.37 bits per parameter; and \textit{Paged Optimizers}, that manage memory spikes during training by utilizing NVIDIA unified memory for automatic CPU-GPU memory transfers.
This combination enables fine-tuning of 65B parameter models on a single 48GB GPU while preserving full 16-bit performance. In practice, QLoRA achieves comparable results to full fine-tuning while requiring only 0.1\% of the trainable parameters, demonstrating particular effectiveness in low-data regimes and better extrapolation to unseen topics.

\textbf{DoRA} \cite{liu2024dora} also uses low-rank matrix decomposition, but its novelty lies in decomposing the pre-trained weight matrix $W$ into magnitude $m$ and direction $V$ components:

\begin{equation}
W = m \frac{V}{\|V\|_c}
\end{equation}

\noindent
where $\|\cdot\|_c$ denotes the vector-wise norm across each column.\\
It then applies LoRA specifically to the directional component:

\begin{equation}
W' = m \frac{W_0 + BA}{\|W_0 + BA\|_c}
\end{equation}
where $W_0$ is the pre-trained weight, and $BA$ represents the low-rank update.
This decomposition approach is inspired by Weight Normalization techniques and aims to simplify the learning task for the low-rank updates.
By doing so, DoRA seeks to enhance the learning capacity and stability of LoRA, potentially bridging the performance gap between PEFT methods and full fine-tuning.
Importantly, \cite{liu2024dora} maintains the inference efficiency of LoRA, introducing no additional latency during deployment. 

On the other hand, \textbf{VB-LoRA} \cite{li2024vb} replaces traditional per-layer adapters with a global pool of reusable vectors, called a vector bank. Rather than learning unique low-rank matrices for each layer, VB-LoRA dynamically constructs them by selecting and combining vectors from this shared bank using a differentiable top-k selection mechanism. This design allows for extreme parameter sharing across the model, significantly reducing memory and storage overhead.

\textbf{VeRA} \cite{kopiczko2024veravectorbasedrandommatrix} is another reparameterization PEFT method for large language models that significantly reduces the number of trainable parameters compared to LoRA while maintaining comparable performance. It uses a single pair of shared random matrices across all layers and learns small scaling vectors for adaptation.
Unlike LoRA, VeRA uses frozen random matrices $A$ and $B$ shared across layers, with trainable scaling vectors $b$ and $d$:

\begin{equation}
    h = W_0x + \Delta Wx = W_0x + \Lambda_bB\Lambda_dAx
\end{equation}
\noindent
where $\Lambda_b$ and $\Lambda_d$ are diagonal matrices formed from vectors $b$ and $d$. Here, $B \in \mathbb{R}^{m \times r}$ and $A \in \mathbb{R}^{r \times n}$ are not required to be low-rank since they remain static. The method employs two initialization strategies:
\begin{itemize}
    \item \textbf{Shared Matrices}: The frozen matrices $A$ and $B$ use Kaiming initialization to maintain consistent variance across all ranks
    \item \textbf{Scaling Vectors}: Vector $b$ is initialized to zeros, while vector $d$ is initialized with a single non-zero value across all elements
\end{itemize}

\begin{table}[htbp]
\centering
\caption{Parameter efficiency comparison of PEFT methods across different pre-trained model architectures. The table shows the number of trainable parameters and the percentage of total parameters updated for each method when applied to various foundation models\\}
\label{tab:peft_comparison}

\begin{adjustbox}{max width=\textwidth}
\begin{tabular}{c|c|c|c}
\hline
\textbf{Model (Total Params)} & \textbf{Method} & \textbf{Trainable Params} & \textbf{\% Trainable} \\
\hline
\multirow{8}{*}{\shortstack{BERT Base Uncased\\[1.5mm](109B)}}
 & Prompt \cite{lester2021power} & 15,360 & 0.01\% \\
 & Prefix \cite{li2021prefix} & 368,640 & 0.34\% \\
 & LoRA \cite{hu2022lora} &  443,906 & 0.41\% \\
 & VeRA \cite{kopiczko2024veravectorbasedrandommatrix} & 57,986 & 0.05\% \\
 & DoRA \cite{liu2024dora} & 471,554 & 0.43\% \\
 & VB-LoRA \cite{li2024vb} & 509,442 & 0.47\% \\
 & QLoRA \cite{dettmers2023qlora}& 443,906&0.40\%\\
 & BitFit \cite{zaken2021bitfit}& 102,914 & 0.09\%  \\
\hline
\multirow{8}{*}{\shortstack{ALBERT Base V2\\[1.5mm](11B)}}
 & Prompt \cite{lester2021power} & 15,360 & 0.13\% \\
 & Prefix \cite{li2021prefix} & 368,640 & 3.15\% \\
 & LoRA \cite{hu2022lora} & 38,402 & 0.33\% \\
 & VeRA \cite{kopiczko2024veravectorbasedrandommatrix} & 6,242 & 0.05\% \\
 & DoRA \cite{liu2024dora} & 40,706 & 0.35\% \\
 & VB-LoRA \cite{li2024vb} & 103,938 & 0.89\% \\
 & QLoRA \cite{dettmers2023qlora}& 38,402 & 0.32\%\\
 & BitFit \cite{zaken2021bitfit}& 10,114 & 0.09\%\\
\hline

\multirow{8}{*}{\shortstack{RoBERTa Base\\[1.5mm](124B)}}
 & Prompt \cite{lester2021power} & 15,360 & 0.01\% \\
 & Prefix \cite{li2021prefix} & 368,640 & 0.30\% \\
 & LoRA \cite{hu2022lora} & 1,034,498 & 0.83\% \\
 & VeRA \cite{kopiczko2024veravectorbasedrandommatrix} & 648,578 & 0.52\% \\
 & DoRA \cite{liu2024dora} & 1,062,146 & 0.85\% \\
 & VB-LoRA \cite{li2024vb} & 1,100,034 & 0.88\% \\
 & QLoRA \cite{dettmers2023qlora}& 1,034,498 & 0.82\%\\
 & BitFit \cite{zaken2021bitfit}& 102,914 & 0.08\%\\
 \hline
 
\multirow{12}{*}{\shortstack{ViT B16\\[1.5mm](85B)}}
 & Prompt \cite{lester2021power} &  15,360 & 0.01\% \\
 & Prefix \cite{li2021prefix} & 384,000 & 0.44\% \\
 & LoRA \cite{hu2022lora} & 442,368 & 0.52\% \\
 & VeRA \cite{kopiczko2024veravectorbasedrandommatrix} & 27,936 & 0.03\% \\
 & DoRA\cite{liu2024dora} & 470,016 & 0.55\% \\
 & VB-LoRA\cite{li2024vb} & 802,816 & 0.54 \% \\
 & QLoRA\cite{dettmers2023qlora}& 43,906 & 0.51\%\\
 & Sequential adapter \cite{pmlr-v97-houlsby19a} & 903,744  & 1.05 \% \\
 & Double adapter \cite{pfeiffer-etal-2020-mad}& 451,872  & 0.05 \% \\
 & Parallel adapter \cite{he2022towards} & 451,872 & 0.05 \% \\
 & Adapter+ \cite{steitz2024adaptersstrike} & 461,088& 0.05 \% \\
 & BitFit \cite{zaken2021bitfit}& 102,914 & 0.12\%\\
 \hline
 
\multirow{12}{*}{\shortstack{CLIP ViT B16\\[1.5mm](149B)}}
 & Prompt \cite{lester2021power} & 15,360 & 0.01\% \\
 & Prefix \cite{li2021prefix} & 675,840 & 0.45\% \\
 & LoRA \cite{hu2022lora} &  737,280  & 0.49\% \\
 & VeRA \cite{kopiczko2024veravectorbasedrandommatrix} &  46,656 & 0.03\% \\
 & DoRA \cite{liu2024dora} & 783,360  & 0.52\% \\
 & VB-LoRA \cite{li2024vb} & 802,816 & 0.54 \% \\
 & QLoRA \cite{liu2024dora}& 737,280 & 0.49\%\\
 & Sequential adapter \cite{pmlr-v97-houlsby19a} & 1,309,632   & 0.87 \% \\
 & Double adapter \cite{pfeiffer-etal-2020-mad} & 654,816 & 0.43 \% \\
 & Parallel adapter \cite{he2022towards} & 654,816 & 0.43 \% \\
 & Adapter+ \cite{steitz2024adaptersstrike} & 670,176& 0.44 \% \\
 & BitFit \cite{zaken2021bitfit} & 171,008 & 0.11\%\\

\hline
\end{tabular}
\end{adjustbox}
\end{table}

\subsection{Selective Methods}
The family of selective methods for PEFT includes such techniques that aim to pick and finetune only a small subset of the model parameters, hence reducing computational requirements and memory footprint.
\textbf{Masking} \cite{zhao2020masking} starts from the idea of selecting the weights that are important for the downstream task, instead of finetuning the entire model.
Based on the \textit{lottery ticket hypothesis} \cite{frankle2018lottery}, they propose to select the parameters with a series of learned binary masks, one for each task.
The authors experimented with Transformer-based architectures, such as BERT \cite{devlin2019bert}, RoBERTa \cite{liu2019roberta} and DistilBERT \cite{sanh2019distilbert}, on different NLP tasks, including part-of-speech tagging, named-entity recognition, sequence classification, and reading comprehension.
In this approach, for each weight matrix $\textbf{W}^l$ of the $l$-th transformer block, they randomly initialize a matrix $\textbf{M}^l$, which is then binarized via an element-wise thresholding function. The binary mask is then applied to the weights by Hadamard product:
$$\hat{\textbf{W}}^l := \textbf{W}^l \odot \textbf{M}^l_\text{bin}$$

Through their experiments, they prove that this masking technique achieves comparable results to complete finetuning, while being more parameter-efficient and lightweight.
They also show that 3\% $\sim$ 10\% initial sparsity of $\textbf{M}^l_\text{bin}$ represents a good trade-off between retaining knowledge and flexibility.
Additionally, they demonstrate that, in general, it is reasonable to select all the weights from shallower layers, while learning masks for higher layers to reach optimal results.

\textbf{BitFit} \cite{zaken2021bitfit} is a sparse finetuning method, in which the basic concept is to update only the bias terms, while keeping the rest of the network frozen.
Such approach is very parameter-efficient, since very few weights are modified: for BERT architecture, they amount to less than 0.1\% of the total number of the parameters.
In particular, the authors consider the bias terms of \textit{key, query} and \textit{value} weights of each self-attention head, for each layer of the BERT encoder, plus the bias parameters from the MLP layers on top.
The BitFit approach was evaluated against the GLUE benchmark \cite{wang2018glue}, obtaining comparable results to naive finetuning, but modifying a tiny portion of the weights.
Moreover, they experimented with updating a subset of the bias parameters, achieving only slightly worse results.

\textcolor{red}{
\section{Categorical Analysis of PEFT: Strengths and Structural Limitations}
\label{subseq:peft_analysis}
}

\textcolor{red}{
The previous subsection introduced a categorical overview of parameter-efficient fine-tuning (PEFT) methods, distinguishing between additive, reparameterization, and selective approaches. While all three families have demonstrated strong performance in static adaptation settings, their behavior changes substantially when applied to Continual Learning (CL). In a sequential learning regime, these methods exhibit different strengths and expose distinct structural limitations that directly impact scalability, stability, and plasticity. This subsection analyzes these trade-offs, highlighting how the design choices underlying each PEFT category influence their suitability for long-sequence continual learning.
}

\textcolor{red}{
\subsection{Additive Methods (Adapters and Prompting)}
In the parameter-efficient continual fine-tuning literature, additive methods represent one of the earliest and most intuitive strategies for adapting large pretrained models in a sequential learning setting. Rather than adjusting the backbone directly, these approaches preserve the original weights and introduce new task-specific components that are trained alongside the frozen model. This structural separation naturally aligns with the stability objective in continual learning, by design, previously learned representations are never overwritten, which sharply reduces catastrophic forgetting. From a practitioner’s standpoint, additive techniques provide a clear division between general knowledge and task-specific adaptation, enabling modular reuse and interpretation. At the same time, this separation shifts the core challenge toward managing the growth and complexity of auxiliary parameters as the number of tasks increases, highlighting an emergent capacity bottleneck in long-horizon continual learning streams. Such dynamics underscore the trade-offs inherent in additive approaches between isolation of task knowledge and sustainable scalability in real-world PEFT settings.
}

\textcolor{red}{
\subsubsection{Strengths.}
\begin{itemize}
    \item \textbf{Modular task representation \& Parameter Efficiency}: Adapters, for example, add small bottleneck modules between layers, enabling task-specific adaptation without modifying backbone weights; this often preserves pre-trained knowledge effectively. Both adapters and soft prompts add only a small fraction of parameters (less than 1\% typically), enabling adaptation to many tasks with minimal memory overhead \cite{hu2023llmadaptersadapterfamilyparameterefficient}.
    \item \textbf{Preserving representation space}: Prompt-based tuning such as \cite{li2021prefix} tends to better preserve the pre-trained feature space compared to weight-modifying methods, helping maintain general representations across tasks \cite{kim2024preservingpretrainedrepresentationspace}.
\end{itemize}
}

\textcolor{red}{
\subsubsection{General Category Limitations for Continual Learning.}
\begin{itemize}
    \item \textbf{Inference overhead \& architectural changes:} Adapters introduce additional layers into the model, which can incur slight inference latency compared to weight-merged methods and requires modifying model structure \cite{electronics14183580}.
    \item \textbf{Sensitivity to placement and configuration}: The performance of adapters like \cite{pmlr-v97-houlsby19a} depends critically on where and how they are inserted: suboptimal placement can degrade performance. 
    \item \textbf{Scalability}: 
    Additive methods encounter significant scalability challenges as the number of learning steps $T$ increases.
    Maintaining a separate adapter for each task leads to linear growth in adapter modules that impairs knowledge reuse and escalates parameter count in continual learning scenarios\cite{wang2023self}. The progressive concatenation of soft prompts continuously increases the effective input sequence length as tasks accumulate~\cite{lester2021power, razdaibiedina2023progressive}. 
    These limitations motivate the development of mechanisms that control and reorganize capacity rather than allowing the system to grow unbounded.
\end{itemize}
}

\textcolor{red}{
\subsection{Reparameterization-Based Methods (LoRA,DoRA,VeRA)}
Reparameterization-based methods have emerged as a central class of parameter-efficient adaptation techniques, particularly in the context of large-scale pretrained models where both compute and memory efficiency are at a premium. Unlike additive methods that append new modules, reparameterization approaches express task-specific updates as low-rank factorization components within the existing weight space. In continual learning, this affords a compelling compromise: adaptation occurs without modifying the  backbone architecture, yet the model retains the flexibility to encode nuanced changes for each new task. From a theoretical perspective, low-rank reparameterization implicitly captures the “intrinsic dimension” of task differences, allowing efficient representation of adaptation with very few parameters. However, when sequential adaptation steps are merged into the same underlying weights, interference between tasks can arise as these low-rank subspaces overlap, leading to subtle forms of forgetting. This interaction reveals a deeper tension in reparameterization methods: they can achieve impressive parameter efficiency and inference consistency, yet must contend with weight-space entanglement that becomes more salient as the continual learning stream lengthens.
}

\textcolor{red}{
\subsubsection{Strengths:}
\begin{itemize}
    \item \textbf{Zero inference latency}:Because the weight updates of \cite{hu2022lora,kopiczko2024veravectorbasedrandommatrix,dettmers2023qlora} can be merged back into the backbone at deployment, there is no additional inference cost beyond the base model.
    \item \textbf{High parameter efficiency with competitive performance}: LoRA consistently enables training a tiny fraction of parameters (e.g., \textless  1\%) while achieving performance close to or matching full fine-tuning on many NLP and vision tasks.
\end{itemize}
}

\textcolor{red}{
\subsubsection{General Category Limitations for Continual Learning:}
\begin{itemize}
    \item \textbf{Rank Selection Sensitivity}: LoRA's performance is highly sensitive to the chosen "rank" (\(r\)). Fixed-rank selection can lead to suboptimal resource allocation, too much for simple tasks and too little for complex ones \cite{zhang2025cloracontinuallowrankadaptation,LING2026112086}. This sensitivity is particularly pronounced in continual learning settings, where tasks may vary widely in complexity. Fixed-rank LoRA configurations may over-allocate capacity for simple tasks while under-allocating for tasks with more intricate patterns or domain shifts.
    \item \textbf{Interference}: Without effective merging mechanisms, new task updates can interfere with previous ones \cite{paeedeh2025continualknowledgeconsolidationlora}.
    \item \textbf{Missed Knowledge Transfer}: Naive LoRA approaches that use separate adapters for every task often fail to capture and reuse shared knowledge across related domains \cite{zhang2025cloracontinuallowrankadaptation,paeedeh2025continualknowledgeconsolidationlora}.
    \item  \textbf{Linear parameter and storage growth}: Standard multi-task LoRA implementations typically add a new set of adapter modules for every new task to prevent interference \cite{wei2025online}. This leads to a linear increase in parameter count and storage overhead as the number of tasks grows, which is not sustainable for true lifelong learning.
\end{itemize}            
}

\textcolor{red}{
\subsection{Selective Methods (BitFit and Partial Tuning)}
Selective methods occupy the most constrained end of the parameter-efficient continuum by deliberately restricting adaptation to a limited subset of existing parameters, such as biases or normalization statistics. In the context of continual learning, this restriction reflects an extreme interpretation of parameter efficiency: rather than adding new capacity or restructuring the update space, these methods assume that only a small fraction of parameters need to change to capture new task behavior. This has clear advantages for deployment in resource-constrained environments, where storage and computational budgets are tight and the cost of storing or updating large parameter sets is unacceptable. At the same time, the selective nature of these approaches raises important questions about expressivity: when only a narrow slice of the model is updated, the capacity to adapt to tasks that diverge significantly from pretrained domains may be limited. Moreover, because selective updates occur within the shared backbone rather than isolated modules, the risk of cross-task interference and forgetting persists unless additional mechanisms are introduced. Thus, selective methods highlight a fundamental tension between ultra-efficient adaptation and the ability to sustain robust performance over long continual learning sequences.
}

\textcolor{red}{
\subsubsection{Strengths:}
\begin{itemize}
    \item \textbf{Extreme storage efficiency}: Because only a tiny fraction of parameters (often \(<0.1\%\) for BitFit) are updated, the "task-specific" differences are extremely compact, making them easy to store for long sequences of tasks \cite{zaken2021bitfit}.
    \item \textbf{Reduced training footprint}: Selective methods significantly lower GPU memory requirements during training because gradients only need to be computed and stored for the small subset of active parameters.
    \item \textbf{Simplicity and speed}: They are often faster to implement and iterate upon than complex architectural modifications like dynamic adapters or routing systems.
\end{itemize}
}

\textcolor{red}{
\subsubsection{General Category Limitations for Continual Learning:}
\begin{itemize}
    \item Selective methods often suffer from \textbf{insufficient expressivity}. Modifying only biases, as in BitFit, is rarely sufficient to adapt large-scale models to radically novel data distributions, resulting in lower peak performance compared to more expressive approaches.
    \item Because selective methods directly modify shared parameters, they provide no mechanism for knowledge isolation. The absence of structural barriers between updates makes them highly susceptible to catastrophic forgetting unless combined with external memory-based techniques.
\end{itemize}
}

\section{The Confluence of CL and PEFT}\label{The Confluence of CL and PEFT}
The intersection between PEFT and CL presents a promising avenue for developing more adaptive and efficient AI systems.
The synergy between these approaches addresses the key limitations that each field has individually and opens up new possibilities for scalable, lifelong learning systems, such as:
\begin{itemize}
    \item \textbf{Efficient Task Adaptation}: Traditional CL methods often struggle with the computational cost of adapting to new tasks while preserving knowledge of previous ones. PEFT techniques such as LoRA \cite{hu2022lora} or adapters \cite{pmlr-v97-houlsby19a} can be used to create task-specific parameter updates with minimal overhead. This allows for rapid adaptation to new tasks without the need for extensive retraining or storing separate models for each task \cite{liu2022few}.

    \item \textbf{Mitigating Catastrophic Forgetting}: CL systems face the challenge of catastrophic forgetting when learning new tasks~\cite{YANG2025108226}. By utilizing PEFT methods, the majority of the model's parameters can remain frozen, preserving knowledge of previous tasks. For instance, Hard Attention to the Task \cite{serra2018overcoming} combines attention mechanisms with selective parameter updates, effectively mitigating forgetting while enabling continual learning.

    \item \textbf{Scalable Multi-task Learning}: As the number of tasks grows, traditional CL approaches often struggle with scalability~\cite{YANG2025108226}. PECFT methods can address this issue by allowing a single large model to adapt to numerous tasks with minimal additional parameters. For instance, \cite{Ke2021AchievingFP} proposed CL-Plugins, a method combining adapter-like modules with continual learning, allowing a single model to incrementally learn multiple tasks while sharing a large portion of parameters.

    \item \textbf{Resource-Constrained Environments}: In scenarios with limited computational resources, such as edge devices or mobile applications, the combination of PEFT and CL enables continuous adaptation without the need for extensive hardware. 

\end{itemize}

In the following sections, we provide a detailed overview of the main PECFT methodologies, categorizing them according to the type of PEFT strategy they employ. A visual taxonomy of these methods is presented in Figure \ref{fig:pecft_taxonomy}.

\begin{figure}
    \centering
    \includegraphics[width=\textwidth]{PECFT_CAT3.pdf}
    \caption{
    Taxonomy of Parameter-Efficient Continual Fine-Tuning methods.
    Methods are categorized by the type of PEFT they use---LoRA, prompts, adapters, or unified methods.
    Following the organizational scheme of Section 5, each method is color-coded to indicate the type of approach employed, often spanning multiple PEFT types.
    Papers are labeled with their publication year and arranged chronologically.
    }
    \label{fig:pecft_taxonomy}
\end{figure}

\subsection{Adapter-based PECFT}
\subsubsection{Routing and Selection Mechanisms}
    
In adapter-based CL frameworks, routing mechanisms enable dynamic selection of specialized adapters, facilitating efficient knowledge accumulation and modular model expansion. \textbf{Mixture-of-Experts (MoE)} \cite{yu2024boostingcontinuallearningvisionlanguage} introduces a framework that dynamically expands pre-trained CLIP models. This approach features a Distribution Discriminative Auto-Selector that intelligently routes inputs between the MoE Adapter and the original CLIP model.  This architecture demonstrates remarkable efficiency, reducing parameter training overhead by 60\% while maintaining competitive performance.

\textbf{SEMA} \cite{wang2024selfexpansionpretrainedmodelsmixture} takes a different approach to routing, employing an expandable weighting router that generates weighted combinations of adapter outputs. Its routing mechanism is enhanced by representation descriptors that monitor distribution shifts to trigger selective adapter expansion. This ``soft'' routing strategy, combined with on-demand expansion, enables SEMA to achieve sub-linear growth rates without relying on memory rehearsal.

\textbf{AdaPtive Adapter RouTing} (APART) \cite{qi2024adaptiveadapterroutinglongtailed} presents an innovative solution to Long-Tailed Class-Incremental Learning (LTCIL), addressing the dual challenges of catastrophic forgetting and data imbalance without relying on stored exemplars.
This approach leverages pre-trained models through a dual-pool adapter system: a primary pool for general knowledge retention, and an auxiliary pool specifically designed for minority classes. APART uses an adaptive instance routing mechanism, which dynamically combines information from both pools without fixed thresholds for minority class identification. By freezing the pre-trained model's core parameters and utilizing trainable layer-wise adapters, the method enables effective adaptation while minimizing forgetting.
APART demonstrates how pre-trained models can be effectively leveraged for LTCIL in real-world applications, where data storage and privacy concerns are paramount, with extensive experimental validation confirming its effectiveness across multiple benchmarks. 

\subsubsection{Merging and Fusion Strategies}

\textbf{ATLAS} \cite{li2024atlasadapterbasedmultimodalcontinual}  proposes a two-stage learning paradigm that effectively balances the preservation of existing knowledge with the acquisition of new capabilities by incorporating vector learning method, which intelligently combines information from different adapters based on cosine similarity between tasks. Unlike previous parameter-efficient solutions that created isolated modules for each task, ATLAS minimizes knowledge redundancy while expanding the model's representational capacity.
It employs both multi-modal and uni-modal tasks in upstream continual learning, providing valuable insights into how multi-modal model updates affect performance across different modalities. Its ability to enhance distribution richness and improve generalization capability makes it a promising solution for CL in Vision-Language models.

\textbf{The Expand and Merge} \cite{10650910} framework introduces a parameter-efficient architecture that combines adapter layers with Vision-Language models. The approach is built on two key innovations: first, it employs specially designed adapter layers that expand for new tasks while keeping old knowledge intact through frozen parameters; second, it leverages a pretrained text encoder's fixed embedding space to guide the vision encoder's continual learning process through vision-language pretraining models like CLIP. To effectively manage knowledge integration, the framework implements an adaptive fusion mechanism using scaling weights at different network depths, complemented by a unique parameter merging stage that prevents performance degradation while controlling parameter growth.
This design addresses two common limitations in the field: the parameter bloat, typical of expansion-based approaches, and the over-constraining of new learning found in regularization-based methods.

\subsubsection{Task Specific Strategies}

\textbf{Continuous Adapter} (C-ADA) \cite{gao2024promptlearningcontinualadapter} explores the potential of rehearsal-free continual learning by introducing an extensible parameter Continual Adapter Layer (CAL) that allows better reuse of knowledge by adapting weights for new tasks while preserving old knowledge.
It also employs a Scaling \& Shifting module, that reduces the divergence between the pre-training and downstream datasets by transferring the feature space from pre-trained datasets to downstream dataset.

Similarly, \textbf{Adapter-based Continual Learning} (ACL) \cite{zhang2023adapter} introduces a framework that uses a feature adapter to combat catastrophic forgetting.
It is designed with a task-specific head, which groups all previously learned classes into a single ``\textit{out-of-distribution}'' category, enabling more effective feature discrimination.
The approach also includes a collaborative fine-tuning mechanism, ensuring that the outputs of the different classifiers remain comparable, facilitating accurate head selection during inference.

The \textbf{Continual Adapter Tuning} (CAT) \cite{CHEN2024127423}  tackles Aspect Sentiment Classification (ASC) by combining task-specific adapters with a frozen pre-trained backbone. CAT uses its continual adapter initialization technique for knowledge transfer between tasks and label-aware contrastive learning to jointly optimize features and classifiers. The framework eliminates the need for task IDs during inference by employing a majority voting strategy across adapter paths. 

\subsubsection{Subspace-based Strategies}

\textbf{Expandable Subspace Ensemble} (EASE) \cite{zhou2024expandable} addresses task interference by creating task-specific subspaces through lightweight adapter expansion. Unlike methods that modify shared parameters, EASE trains a distinct adapter for each new task, enabling conflict-free learning where new tasks don't harm previous knowledge. The method concatenates embeddings from all task-specific subspaces to create high-dimensional representations for holistic decision-making.
EASE utilizes a semantic-guided prototype complement strategy that synthesizes old class prototypes in new subspaces without exemplars. It extracts class-wise similarity in co-occurrence spaces and uses this semantic information to reconstruct missing prototypes. During inference, subspace reweighting highlights contributions from matching task-specific adapters.

\subsection{LoRA-Based PECFT}

Recent works have leveraged Low-Rank Adaptation for CL scenarios, offering a parameter-efficient alternative to traditional prompt or adapter-based methods. By utilizing LoRA's inherent ability to create compact task-specific updates, these approaches demonstrate strong performance in sequential learning while maintaining minimal memory overhead. 

\subsubsection{Orthogonal Subspace Methods}
    
\textbf{InfLoRA} \cite{liang2024inflora} is a reparameterization technique that confines weight updates to a carefully designed subspace, eliminating interference between new and old tasks. This subspace-constrained learning improves upon existing approaches that either reuse parameters or randomly expand them for new tasks. The method approximates the gradient space using the input matrix of new tasks, making it particularly effective in CIL scenarios where task IDs are unavailable during inference.

\textbf{Prototype Guided Incremental LoRA (PILoRA)}  \cite{guo2024federated} addresses two key challenges in federated class incremental learning: catastrophic forgetting and data heterogeneity across clients.
The method combines prototype learning with parameter-efficient fine-tuning through an innovative two-pronged approach. First, it introduces incremental LoRA, which mitigates forgetting by constraining different learning stages to orthogonal subspaces, allowing efficient knowledge accumulation through parameter summation during inference. Second, it implements a prototype re-weight module that leverages heuristic information between prototypes and class features to address classifier bias without retraining.
Its effectiveness stems from the unified approach to both continual learning and federated learning challenges, recognizing their shared need for robust feature representations and bias mitigation in classifiers.
While InfLoRA focuses on confining weight updates to carefully designed subspaces to prevent interference between tasks, PILoRA expands on this idea by implementing orthogonal subspaces for different learning stages while adding prototype learning to handle data heterogeneity.

\textbf{SD-LoRA}\cite{wu2025sdlorascalabledecoupledlowrank} introduces a decoupled low-rank adaptation method for class-incremental learning that explicitly separates the learning of direction and magnitude in LoRA updates. After learning the low-rank direction matrices during the first task, SD-LoRA freezes them and optimizes only the task-specific magnitude vectors for subsequent tasks. This design ensures that all tasks share a common representational subspace, reducing interference and eliminating the need for rehearsal or dynamic routing. The approach is fully end-to-end, does not expand the model over time, and supports task-agnostic inference using a single unified adapter.

A slightly different approach comes from \textbf{Dual Low-Rank Adaptation} (DualLoRA) \cite{chen2024duallowrankadaptationcontinual}.
It employs two distinct low-rank adapters that work in concert: an orthogonal adapter, which operates in subspaces perpendicular to previous task features, to preserve old knowledge, and a residual adapter, which learns in task-specific subspaces to enable effective adaptation to new tasks.
DualLoRA is quite efficient in feature subspace extraction, as it utilises Singular Value Decomposition, which eliminates the need for multiple forward passes that burden previous approaches, such as InfLoRA.
During inference, DualLoRA uses its dynamic memory mechanism which modulates the residual adapter's output based on task relevance computed from input samples. This dynamic adjustment not only enhances feature embeddings but also facilitates accurate task identification without requiring explicit task labels.

\textbf{LoRAC-IPC} \cite{ling2025lora}, on the other hand, addresses the challenge of parameter interference in LoRA-based continual learning. The method introduces a two-stage approach: Importance-based Parameter Consolidation (IPC) and Orthogonal LoRA Composition (LoRAC). In the first stage, \textbf{LoRAC-IPC} identifies critical LoRA parameters from earlier tasks, those contributing most to past performance, and freezes them. In the second stage, it ensures that new LoRA updates are orthogonal to frozen directions using QR decomposition, effectively preventing destructive interference. Unlike naive LoRA stacking or merging, this composition maintains compatibility across tasks 

\subsubsection{Merging and Fusion Strategies}

\textbf{Interpolation-based LoRA} \cite{ren2024analyzingreducingcatastrophicforgetting} approaches CL in LLMs through mode connectivity, \textit{i.e.} the observation that different optimization minima can be connected through low-loss valleys.
Rather than viewing catastrophic forgetting as a binary trade-off, it implements a dual-memory architecture: a fast learner that rapidly adapts to new tasks, which provides plasticity, and a slow learner that consolidates long-term knowledge, maintaining stability.
This design is inspired by empirical findings that mode connectivity exists in parameter-efficient fine-tuning of LLMs, and that linear interpolation between task-specific optima can achieve better plasticity-stability trade-offs than previous approaches.

\textbf{MoLE} \cite{wu2024mixtureloraexperts} introduces a hierarchical mixture-of-experts framework for composing multiple LoRA adapters without retraining the base model. Rather than using naive arithmetic merging, MoLE treats each LoRA as an expert and learns layer-wise softmax-normalized gating weights to combine their contributions. Specifically, for each transformer layer, the LoRA updates from all adapters are merged into a single update via a learned weighted sum, where the weights are unique to each layer. This approach allows fine-grained control over how task-specific knowledge is integrated, enabling modular, task-agnostic inference without the need for retraining or access to task IDs. 

Merging-based approach is used in \textbf{Task Arithmetic with LoRA for CL} \cite{chitale2023task}.
Instead of sequential training, which introduces forgetting risks, this approach trains individual LoRA modules for each task separately. It then combines the task vectors using task-specific arithmetic rules before merging them back into the pre-trained Vision Transformer. The method's innovation lies in its efficient use of resources, as it only trains small LoRA modules and leverages a minimal memory buffer of 10 samples per class for final fine-tuning.

\subsubsection{Masking-based Strategies}

CL with \textbf{STack-And-Mask INcremental Adapters} (STAMINA) \cite{Smith2023ContinualDW} enhances text-to-image diffusion models' ability to learn long sequences of concepts without forgetting. The method combines LoRA with two key technical innovations: hard-attention masks parameterized by low-rank MLPs using Gumbel softmax, and learnable MLP tokens that replace traditional custom token embeddings, with all trainable parameters capable of being folded back into the base model after training, eliminating inference overhead.
It is also able to maintain plasticity thanks to its sparse adaptation approach.

\subsection{Prompt-based PECFT}
\subsubsection{Prompt Pool-based Strategies}
\textbf{Learning to Prompt} (L2P) \cite{wang2022learning} is one of the first works that applies prompting to CL.
In this work, a prompt is trained for each task on a pre-trained network and subsequently stored in a key-value \textit{prompt pool}, where the key is learnable.
To select the most adequate prompts at inference time for a given input, the authors introduce a deterministic query function, \textit{i.e.} the pre-trained network, to extract features from the input.
Prompts are then selected based on the similarity between the input query and prompt keys.
L2P was evaluated against the main CL strategies in class-incremental, domain-incremental, and task-agnostic scenarios.

Similarly, \textbf{DualPrompt} \cite{wang2022dualprompt} builds on the prompt pool concept by introducing two types of prompts: \textit{general prompts}, for common features across tasks, and \textit{expert prompts}, for task-specific instructions.
These prompts are attached to two distinct groups of contiguous attention layers, selected through heuristic search.
DualPrompt leverages the L2P workflow: each expert prompt is associated with a task-specific key that is learned to match the input features; at inference time, a query function on the test sample is used to retrieve the best e-prompt.

A slight variation from L2P comes from \textbf{S-Prompts} \cite{wang2022s}.
This approach is specifically applied to Domain-Incremental Learning scenarios, without the need to keep a buffer of samples from previous tasks.
In L2P and similar works, the prompts are shared among tasks, meaning that the new knowledge shares the same feature space of the old tasks, limiting the learning capacity and possibly inducing interference.
Conversely, S-Prompts instantiate a new prompt for each domain, so that each prompt can obtain optimal performance on its task independently of the other prompts.
At inference time, K-NN is employed to find the closest domain to a given test sample, and the corresponding prompt is prepended to the input to perform classification.

\textbf{Language Guidance for Prompt-based Continual Learning} (LGCL) \cite{khan2023introducing} introduces a method for selecting prompts using semantic information derived from natural language. It uses a pre-trained text encoder to generate task- and class-level embeddings. Task embeddings are matched to prompt keys via cosine similarity to guide prompt selection, while class embeddings are injected into the output features for classification.

\subsubsection{Prompt Composition and Sequential Learning}

One drawback of prompt-based methods we have considered up to this point lies in the fixed dimensions of both the prompt pool and the prompts themselves.
This limits their capability to scale, making it difficult to increase the learning capacity when needed.
\textbf{CODA-Prompt} \cite{smith2023coda} overcomes this issue by defining a set of \textit{prompt components}: for each input sample, the components are merged via weighted summation, forming a so-called \textit{decomposed} prompt.
By doing so, the method can automatically adjust the prompting capacity depending on the complexity of the task at hand.
Furthermore, the authors introduced an attention mechanism on the input query, which allows focusing on relevant features only, reducing also the input dimensionality.
CODA-Prompt is optimized in an end-to-end fashion, using directly the classification loss, which helps increase performance.

\textbf{Progressive Prompts} \cite{razdaibiedina2023progressive} offers a more direct approach.
In this work, a new prompt is learned for each new task, and it is then concatenated with the previously learned ones. Each prompt $P_k$ is trained only during the corresponding task $T_k$, while it is kept frozen during subsequent tasks.
The prompts' concatenation is then prepended to the input embeddings.
This approach was evaluated on two CL benchmarks for text classification using two widely used language models.
The results demonstrate its effectiveness in improving performance and mitigating catastrophic forgetting by facilitating greater forward transfer across tasks.

Following the idea of prompts concatenation, \textbf{Prompt of Prompts} (POP) \cite{hu2023pop} proposes an approach involving two sets of prompts.
First, one or more prompts are learned for each task, with the objective of distinguishing between classes in the task.
Such prompts are then kept frozen for subsequent tasks, but are still given in input to the model; in this way, the new prompts are forced to learn only the new information from the current task, without duplicating knowledge present in other prompts.
Additionally, this method uses a second set of prompts, called \textit{Prompt of Prompts}, to combine the representations from different tasks.
While the task prompts are frozen after training on the corresponding task, the \textit{Prompt of Prompts} set is learned continually, so as to integrate and update the features across all the tasks.

\subsubsection{Orthogonal Subspace Approaches}

\textbf{Prompt Gradient Projection} (PGP) approach \cite{qiao2023prompt} represents the first work that analyzes anti-forgetting mechanisms integrated with prompt tuning techniques.
In particular, the authors employ the Gradient Projection Method \cite{saha2021gradientprojectionmemorycontinual}, which demonstrates theoretically that forgetting is mitigated if the weights are updated in the orthogonal direction to the subspace spanned by the previous features.

\textbf{Visual Prompt Tuning} (VPT) \cite{NEURIPS2024_0f06be00} builds on prompt-based continual learning by addressing the interference introduced during prompt updates in vision transformers. The authors hypothesize that naively updating prompts across tasks leads to inconsistent gradient directions and degrades performance. To counter this, VPT projects the prompt update into the null space of previously learned prompt gradients, ensuring that new learning has minimal impact on prior knowledge. Unlike earlier works, VPT‑NSP takes into account ViT-specific components such as LayerNorm and self-attention when computing the projection basis, making it better aligned with the model’s internal representations. Experimental results show that enforcing gradient orthogonality at the prompt level is effective in reducing forgetting in ViT-based continual learning.

\subsection{Unified Frameworks}
Unlike most PEFT-based continual learning methods that focus on a single technique, such as adapters, prompts, or LoRA, some recent works propose unified frameworks that offer flexibility in applying various PEFT strategies within a continual learning setting. These frameworks aim to decouple the continual learning strategy from the specific choice of PEFT module, enabling broader applicability and easier experimentation.

\textbf{APER} (AdaPt and mERge) \cite{10.1007/s11263-024-02218-0} presents a unified framework that can be orthogonally combined with any parameter-efficient tuning method to address class-incremental learning. The method operates through a two-stage protocol: first adapting the pre-trained model using any PEFT technique (VPT-Deep/Shallow, Scale \& Shift, Adapter modules, or full fine-tuning) on only the first incremental task, then concatenating the embeddings from both the adapted model and the original frozen PTM to create unified representations. This concatenated embedding function remains frozen throughout subsequent learning stages, with new classes handled through prototype-based classifiers that leverage both the task-specific adaptivity of the fine-tuned model and the broad generalizability of the original PTM. A key insight from this work is that a simple baseline (SimpleCIL) using only frozen PTM features with prototype classifiers can outperform state-of-the-art methods like L2P and DualPrompt by significant margins, demonstrating the inherent power of pre-trained representations for continual learning. The framework's universality is evidenced by its compatibility with diverse architectures (ViTs and CNNs) and multiple PEFT approaches, while maintaining parameter efficiency and avoiding catastrophic forgetting through its adapt-once, merge-and-freeze strategy that balances the fundamental trade-off between generalizability and adaptivity in the pre-trained model era.

\textbf{LAE} (Learning-Accumulation-Ensemble) \cite{gao2023unified} proposes a unified continual learning framework that can transform any parameter-efficient tuning method into an effective continual learning approach. The framework positions prompting as merely one instantiation of PEFT and extends the concept to work with general parameter-efficient methods including Adapter, LoRA, and Prefix tuning. LAE operates through three key components: (1) \textit{Learning} with calibrated speed to align the adaptation rates of different PEFT modules, addressing the issue that different PEFT methods vary in their knowledge acquisition speed which can lead to overfitting or insufficient plasticity; (2) \textit{Accumulation} of multi-task knowledge where task-specific information learned by online PEFT modules is progressively accumulated into offline PET modules through sophisticated knowledge transfer mechanisms; and (3) \textit{Ensemble} of two expert models constructed with online and offline PEFT modules that work collaboratively during inference to balance stability and plasticity. The framework's key contribution lies in its ability to reshape any given PEFT method into a competitive memory-free continual learning approach while maintaining the computational efficiency advantages of parameter-efficient tuning. By providing theoretical foundations for why different PEFT methods require speed calibration and demonstrating superior performance across CIFAR100 and ImageNet-R benchmarks, LAE establishes a general paradigm that unifies the benefits of various parameter-efficient approaches under a single continual learning framework, making it broadly applicable across different model architectures and tuning strategies.
{
While both \textbf{APER} and \textbf{LAE} employ techniques for accumulating knowledge efficiently, their core design philosophies for managing knowledge transfer differ significantly. APER appears to focus on a one-time adaptation (on the first task) followed by merging and freezing, making it highly suitable for scenarios where the feature extraction requirements remain relatively fixed across subsequent tasks. In contrast, LAE, with its 'Learn-Accumulate-Ensemble' mechanism, offers a more dynamic, ensemble-based knowledge accumulation model, which may be better suited for long learning sequences with greater inter-task variance as it maintains plasticity by learning a new low-rank matrix for each task.
}



\subsection{Performance Analysis and Trade-offs}
\label{sec:metrics}
The problem of correctly evaluating the performance of CL algorithms has been present since the early developments of the field.
It is clear that accuracy alone is insufficient to comprehensively assess these methods in CL scenarios: traditional ``\textit{offline}'' learning models typically achieve higher accuracy than CL solutions.
However, dynamic environments introduce additional performance considerations, making it necessary to balance trade-offs among the multiple evaluation metrics.
This is particularly evident in the context of PECFT methods.
Indeed, we have mentioned multiple times throughout this work that the principal concern about PTMs lies in their training and inference inefficiency, as well as the large resources required for fine-tuning.
This means that, when dealing with very large architectures, pure accuracy cannot be the only metric employed to evaluate such models.

Firstly, CL scenarios usually involve multiple learning tasks, each with its own training and a test sets.
At first glance, the accuracy on the final task might seem like a meaningful indicator of overall model capabilities, since it represents the end of the training phase.
However, a core principle of CL is the ability to retain knowledge across the entire learning experience.
As such, more comprehensive accuracy metrics are required to capture this aspect effectively.
Formally, we define the \textbf{Average Accuracy} \cite{chaudhry2019efficientlifelonglearningagem} as:
\begin{equation}
A_i = \frac{1}{N} \sum_{j=1}^N a_{N,j}
\end{equation}
\noindent
where $a_{i,j}$ is the accuracy on task $j$, with the model trained continually from task 1 to task $N$.

Apart from pure accuracy, the CL fields principally study approaches to diminish the catastrophic forgetting phenomenon; hence it comes straightforwardly that it represents an important measure to keep into consideration.\\
\textbf{Average Forgetting} \cite{chaudhry2018riemannian} represents how much the model ``\textit{forgets}'' about previous tasks.
It is formally defined as:
\begin{equation}
\begin{aligned}
F_i &= \frac{1}{i} \sum_{j=1}^{i-1} f_{i,j}\\
\text{where } f_{k,j} = \max&_{l \in \{ 1, \dots, k-1 \}} (a_{l,j} - a_{k,j}), \quad \forall j < k
\end{aligned}
\end{equation}

Given the sequential nature of the learning scenarios, it is useful to measure what is the influence of learning multiple tasks, \textit{i.e.} how much training on a task affects the performance on other tasks.
This is particularly interesting because it gives practical insights on how fast the model learns through time.
For this reason, forward transfer and backward transfer \cite{lopez2017gradient, diaz2018don} are introduced.\\
\textbf{Forward Transfer} (FWT) measures the influence a task $t$ has when training on a future task $k$; a positive FWT indicates that the model is able to exploit previously acquired knowledge during the subsequent learning sessions.
It is defined as:
\begin{equation}
FWT = \frac{\sum_{i<j} a_{i,j}}{ \frac{N(N-1)}{2} } 
\end{equation}
\noindent
Similarly, \textbf{Backward Transfer} (BWT) indicates the influence of a task $t$ on a previously learned task $k$; a positive BWT means the agent is able to not degrade and improve performance throughout its lifetime.
\begin{equation}
    BWT = \frac{\sum_{i=2}^N \sum_{j=1}^{i-1} (a_{i,j} - a_{j,j}) }{ \frac{N(N-1)}{2} }
\end{equation}

Other meaningful metrics to keep track of are the ones related to the model efficiency.
An absolute measure of the model size is given by the \textbf{Number of trainable parameters} \cite{hu2022lora}, \textit{i.e.} the amount of parameters introduced by a PECFT method.
This represents a straightforward yet effective measure of the computational requirements of a given approach, and a simple way to compare different techniques.

On the same line, the relative growth of the number of parameters over time is equally important.
We define the \textbf{Model Size Efficiency} (MS) \cite{diaz2018don} as:
\begin{equation}
MS = \min(1, \frac{\sum_{i=1}^N \frac{Mem(\theta_1)}{Mem(\theta_i)} }{N})
\end{equation}
\noindent
where we indicate with $Mem(\theta_i)$ the model memory size in terms of the parameter $\theta$ at the task $i$.

\subsection{Comparative Performance Evaluation}
The current AI landscape is filled with a vast number of benchmarks and evaluation metrics, and individual methods often rely on different experimental settings.
On top of that, a wide set of variables, such as the number of training epochs, the learning rate and PEFT-specific parameters, \textit{e.g.} LoRA rank or prompt size, can significantly influence performance.
{This variability substantially complicates fair comparisons across models, limiting the ability to determine not only which methods perform best under specific conditions, but also the underlying reasons for their relative effectiveness.
To address this issue, the following section presents an experimental analysis of representative PECFT methods, with the goal of providing performance-oriented insights to guide future research.}

{
We focus on Computer Vision tasks, as this domain represents one of the most extensively studied research areas and is characterized by higher intrinsic complexity compared to many NLP benchmarks.
In fact, image data often exhibit substantial variability across tasks, leading to pronounced distribution shifts that pose significant challenges for CL methods.
In contrast, NLP tasks share a common linguistic structure, which reduces inter-task distribution shift and generally results in less severe forgetting.
}

{
We select three widely used datasets as benchmarks:
\begin{itemize}
    \item \textbf{CIFAR-100} \cite{krizhevsky2009learning}, which contains 60{,}000 images spanning 100 classes.
    \item \textbf{CUB-200} \cite{wah2011caltech}, comprising 11{,}788 images of birds distributed across 200 fine-grained categories.
    \item \textbf{TinyImageNet} \cite{le2015tiny}, a reduced-scale variant of ImageNet \cite{deng2009imagenet}, including 100{,}000 images across 200 classes.
\end{itemize}
}

{
We report two standard CL metrics, as described in Section \ref{sec:metrics}: average accuracy and average forgetting.
Together, these metrics allow the assessment of both a model’s learning capability---namely, its ability to acquire and perform well on individual tasks---and its capacity to retain knowledge across tasks.
}

{
We conduct this evaluation across a set of representative PECFT methods.
As the base model, we employ a ViT-B/16 backbone pre-trained on ImageNet.
For L2P, DualPrompt, and CODA-Prompt, we use the implementations provided by the \texttt{mammoth} library \cite{boschini2022class, buzzega2020dark}.
All remaining methods are evaluated using the official implementations released by their respective authors.
}

\subsubsection{Accuracy Performance Analysis}
The comparative analysis of PECFT methods reveals a complex landscape where no single approach achieves universal superiority, with performance being highly dependent on both dataset characteristics and task sequence length. While this evaluation focuses on the most widely adopted PECFT approaches rather than exhaustively covering all variants, the selected methods represent the core paradigms that have gained significant traction in the field, providing meaningful insights into fundamental trade-offs.
The results of our evaluation of the different methods are available in Table \ref{tab:accuracy}.

\begin{table}
\centering
\begin{adjustbox}{max width=\textwidth}
\begin{tabular}{l l l l lll}\toprule
Method & \multicolumn{2}{c}{CIFAR-100} & \multicolumn{2}{c}{CUB-200} & \multicolumn{2}{c}{Tiny ImageNet}\\\midrule

 & \textbf{10 Tasks} & \textbf{20 Tasks} & \textbf{10 Tasks} & \textbf{20 Tasks}  & \textbf{10 Tasks} & \textbf{20 Tasks}\\
\midrule
L2P & 80.01 ± 0.93 & 72.48 ± 0.33 & 64.32 ± 0.18 & 56.62 ± 0.77  & 80.24 ± 0.21&75.44 ± 0.87\\
DualPrompt & 80.45 ± 0.33 & 77.83 ± 0.40 & 69.29 ± 2.73 & 57.77 ± 3.58  & 83.32 ± 0.27&78.77 ± 0.01\\
CODA-Prompt & 83.65 ± 1.34 & 74.18 ± 1.03 & 73.23 ± 0.21 & 60.69 ± 1.97  & \textbf{89.33 ± 0.40} & 83.88 ± 0.27\\
InfLoRA & 84.64 ± 0.41 & 78.14 ± 0.41 & 73.42 ± 1.06 & 63.61 ± 0.37  & 83.29 ± 0.18 & 80.21 ± 0.55\\
SD-LoRA & 86.02 ± 0.24 & 82.08 ± 1.92 & 72.07 ± 3.77 & 68.36 ± 3.06  & 88.76 ± 0.65&85.32 ± 0.82\\
LoRAC-IPC & \textbf{87.79 ± 0.19} & \textbf{84.59 ± 0.51} & 81.24 ± 0.57 & 72.44 ± 1.91 & 84.84 ± 0.27 & 83.43 ± 0.23\\
SEMA & \underline{86.41 ± 0.55} & 81.88 ± 0.92 & 74.19 ± 0.96 & 65.15 ± 0.95  & \underline{89.07 ± 0.69} & \textbf{87.27 ± 0.38}\\
APER (adapter) & 84.16 ± 0.14 & 82.06 ± 0.23 & \textbf{86.31 ± 0.01} & \textbf{86.12 ± 0.04} & 76.84 ± 0.54 & 70.31 ± 0.20\\
EASE & 84.29 ± 0.54 & 82.51 ± 0.12 & \underline{81.60 ± 0.28} & \underline{81.29 ± 0.07} & 88.39 ± 0.08 & \underline{85.85 ± 0.13}\\
LAE & 84.84 ± 0.13 & \underline{83.52 ± 0.12} & 79.77 ± 0.45 & 73.74 ± 0.43 & 85.58 ± 0.15 & 83.52 ± 0.12\\
\bottomrule

\end{tabular}
\end{adjustbox}
\caption{Average Accuracy (\%, $\uparrow$) of PECFT methods after seeing all tasks. Each method is assessed across multiple datasets and learning scenarios. The best performance is highlighted in \textbf{bold}, while the second-best result is \underline{underlined}.}
\label{tab:accuracy}
\end{table}

\paragraph{Adapter-based methods}
Adapter-based methods demonstrate consistent performance across the different benchmarks, with APER achieving exceptional results on fine-grained classification tasks (86.31\% on CUB-200 with 10 tasks), showing also remarkable scalability when increasing the number of tasks (-0.19\% accuracy).
EASE shows competitive performance, obtaining the second best results in multiple scenarios, while demonstrating good scalability in longer scenarios.
SEMA has strong performance on broader classification tasks (86.41\% on CIFAR-100, 89.07\% on Tiny-ImageNet), obtaining instead slightly worse results on CUB-200. It also suffers severe scalability issues, especially on CUB-200, where it has a 9.04\% drop in performance when going from 10 to 20 tasks scenarios.
On average, adapter-based models obtain the highest accuracy, which is expected since they typically require the higher number of trainable parameters, as seen in Table \ref{tab:peft_comparison}.



\paragraph{LoRA-based methods}
LoRA approaches perform competitively, with LoRAC-IPC achieving the highest scores on CIFAR-100. Together with SD-LoRA, it shows consistent cross-dataset performance, with moderate but predictable scalability, 3.4-3.9\% degradation on longer sequences. 
On the other hand, InfLoRA's subspace constraints show concerning scalability issues on complex datasets, despite superior forgetting metrics.
Overall, LoRA-based methods are reliable choices for diverse scenarios, also considering their increased efficiency compared to adapter-based architectures.


\paragraph{Prompt-based methods}
Traditional prompt-based methods face fundamental limitations, primarily due to the limited learning capacity resulting from the small number of additional trainable parameters they involve.
Even more involved variants, like CODA-Prompt, exhibit poor scalability, showing accuracy drops ranging from 9.47\% to 12.54\% as the number of tasks increases, and consistently lower absolute performance, lagging 3-6\% behind other PECFT alternatives.
In contrast, simpler methods like DualPrompt surprisingly demonstrate greater stability across different domains.
In summary, prompt-based methods represent the lightest and most efficient PECFT alternatives, showing nonetheless inferior performance when compared to other approaches.


\paragraph{Unified frameworks}
Unified frameworks like LAE show promising scalability, achieving only a 1.32\% drop on CIFAR-100, indicating that these meta frameworks can effectively leverage the strengths of their underlying PEFT components.
However, their evaluation within an adapter-based setting limits our understanding on how well they generalize across different PEFT strategies.

\vspace{3mm}
These results suggest that practitioners should prioritize adapter methods for peak performance on specific domains, LoRA methods for consistent multi-domain deployment, and consider unified frameworks as promising approaches for combining multiple PEFT strengths, while carefully weighing the scalability-performance trade-offs inherent in each approach.\includecomment{as the 10-to-20 task scaling reveals that method stability often matters more than initial performance in long-term continual learning scenarios.} 
{Finally, while the empirical analysis presented here focuses on sequences up to 20 tasks, it is crucial to consider the theoretical implications for extremely long learning sequences, e.g., hundreds or thousands of tasks. In these highly demanding scenarios, parameter management becomes paramount. Parameter-isolation strategies (such as L2P or task-specific LoRA modules) scale efficiently by adding new, lightweight parameters decoupled from the total number of tasks, preventing catastrophic parameter growth. Crucially, parameter-merging or reparameterization methods, such as those explored in recent work like Hierarchical Adapter Merging \cite{coleman2025hamhierarchicaladaptermerging}, also offer compelling solutions. These approaches dynamically compose or compress a fixed pool of parameters, thus offering strong theoretical scalability by mitigating the issue of model size growth, suggesting superior resilience to interference in the limit of very long sequential task flows.}

\begin{table}
    \centering
    \resizebox{50mm}{!}{
    \begin{tabular}{ccc}\toprule
        Method & Average Forgetting\\\midrule
        L2P & 10.61 ± 0.37\\
        DualPrompt & 6.43 ± 0.14\\
        CODA-Prompt & \underline{4.98 ± 0.52}\\
        InfLoRA & \textbf{4.61 ± 0.44}\\
        SD-LoRA & 6.19 ± 1.09 \\
        LoRAC-IPC & 5.83 ± 0.38\\
        SEMA & 6.40 ± 0.87\\
        APER \tiny{adapter} & 6.93 ± 0.14\\
        EASE & 6.80 ± 0.11\\
        LAE & 6.18 ± 0.19 \\
 \bottomrule
    \end{tabular}}
    \caption{Average Forgetting(\%, $\downarrow$) of different PECFT methods, assessed on CIFAR100 with 10 tasks. The best performance is highlighted in \textbf{bold}, while the second-best result is \underline{underlined}.}
    \label{tab:forgetting}
\end{table}

\subsubsection{Forgetting Performance Analysis}

{
The catastrophic forgetting analysis reported in Table \ref{tab:forgetting} indicates heterogeneous forgetting behaviors across methods, which do not directly correlate with overall performance metrics.
This observation complicates the interpretation of method effectiveness when evaluated solely on accuracy-based metrics.}

{
Among LoRA-based approaches, InfLoRA exhibits the lowest forgetting value (4.61\%), consistent with its subspace-constrained update mechanism, which limits interference with previously learned parameters.
LoRAC-IPC shows intermediate forgetting levels, suggesting that its two-stage training procedure partially mitigates parameter interference.
SD-LoRA presents comparable but slightly higher forgetting, indicating that decoupled update strategies alone may be insufficient to achieve the same level of stability as subspace-constrained formulations.
}

{
Adapter-based methods display relatively uniform forgetting behavior. Specifically, SEMA, EASE and APER report forgetting values in the 6--7\% range, despite differences in their final performance.
This pattern suggests that adapter-based architectures may provide a similar degree of stability across methods, while not significantly reducing forgetting beyond this baseline.
}

{
Prompt-based methods exhibit a wider range of forgetting outcomes.
CODA-Prompt achieves a forgetting value of 4.98\%, comparable to the lowest values observed among LoRA-based methods, whereas L2P records substantially higher forgetting (10.61\%).
DualPrompt yields intermediate forgetting, indicating that design choices within prompt-based frameworks have a measurable impact on stability.
}

{
Overall, the results highlights a weak coupling between performance and forgetting across methods.
InfLoRA combines low forgetting with competitive performance, whereas some high-performing methods, such as SEMA, achieve strong accuracy while exhibiting higher forgetting.
Conversely, methods such as L2P demonstrate both elevated forgetting and lower performance.
These findings suggest that trade-offs between performance and stability remain prevalent, and that approaches based on constrained parameter updates may offer a more balanced solution for long-term CL scenarios.
}

{
These empirical observations motivate a closer examination of the design and architectural choices adopted by PECFT methods, as such choices play a central role in shaping the observed trade-offs between performance and forgetting.
}

\subsection{Core Design Choices of PECFT}
{
PECFT methods adopt a range of architectural and design choices, each introducing distinct advantages and trade-offs.
Understanding how these choices influence CL performance is therefore essential when applying PEFT techniques.
Rather than treating PECFT methods as isolated approaches, we distill a set of core design principles informed by both empirical evidence from our comparative analysis and theoretical insights from the literature.
}

\subsubsection{Parameter Separation}
When working with PECFT, a key design choice is deciding which parameters should remain frozen and which ones should be allowed to change for new tasks. This decision is crucial to how the methods learn new tasks and also maintain old knowledge.\\
{
One illustrative example is freezing the backbone.
Most PECFT methods that achieve strong performance retain the pre-trained model parameters while introducing task-specific adaptations via additional modules.
This approach leverages the general-purpose representations encoded in pre-trained models, which capture fundamental patterns that are broadly useful across domains.
Modifying these representations during continual learning can lead to the erosion of this general knowledge and, consequently, catastrophic forgetting.
By keeping the backbone frozen, PECFT methods preserve these foundational representations while allowing task-specific components to acquire new knowledge through dedicated parameters.
}

Building on the freezing of the backbone, the next critical decision is how to introduce task-specific adaptations.
{
In fact, PECFT methods differ fundamentally in the way they introduce learning capacity for new tasks on top of the pre-trained backbone.  
Specifically, PECFT methods augment the model in three distinct ways:
}

\begin{itemize}
\item \textbf{Modular Separation:}
{
Adds dedicated task-specific modules on top of a frozen backbone.
This approach allows clear modularity between tasks and preserves the pre-trained representations.
It typically requires careful design of routing or selection mechanisms to ensure the correct module is applied for each task.
}

\item \textbf{Subspace Separation:}
{
Restricts parameter updates to a low-dimensional subspace of the original model parameters.
This method achieves parameter efficiency and reduces interference with previously learned knowledge, though some residual interference between tasks may still occur due to overlapping subspaces.
}

\item \textbf{Input Space Separation:}
{
Modifies the input representation to incorporate task-specific information while keeping all model parameters unchanged.
This enables learning new tasks without altering the backbone, but the effectiveness strongly depends on the richness of the input transformations and their ability to encode task-relevant features.
}
\end{itemize}

\begin{figure}
    \centering
    \includegraphics[width=1\linewidth]{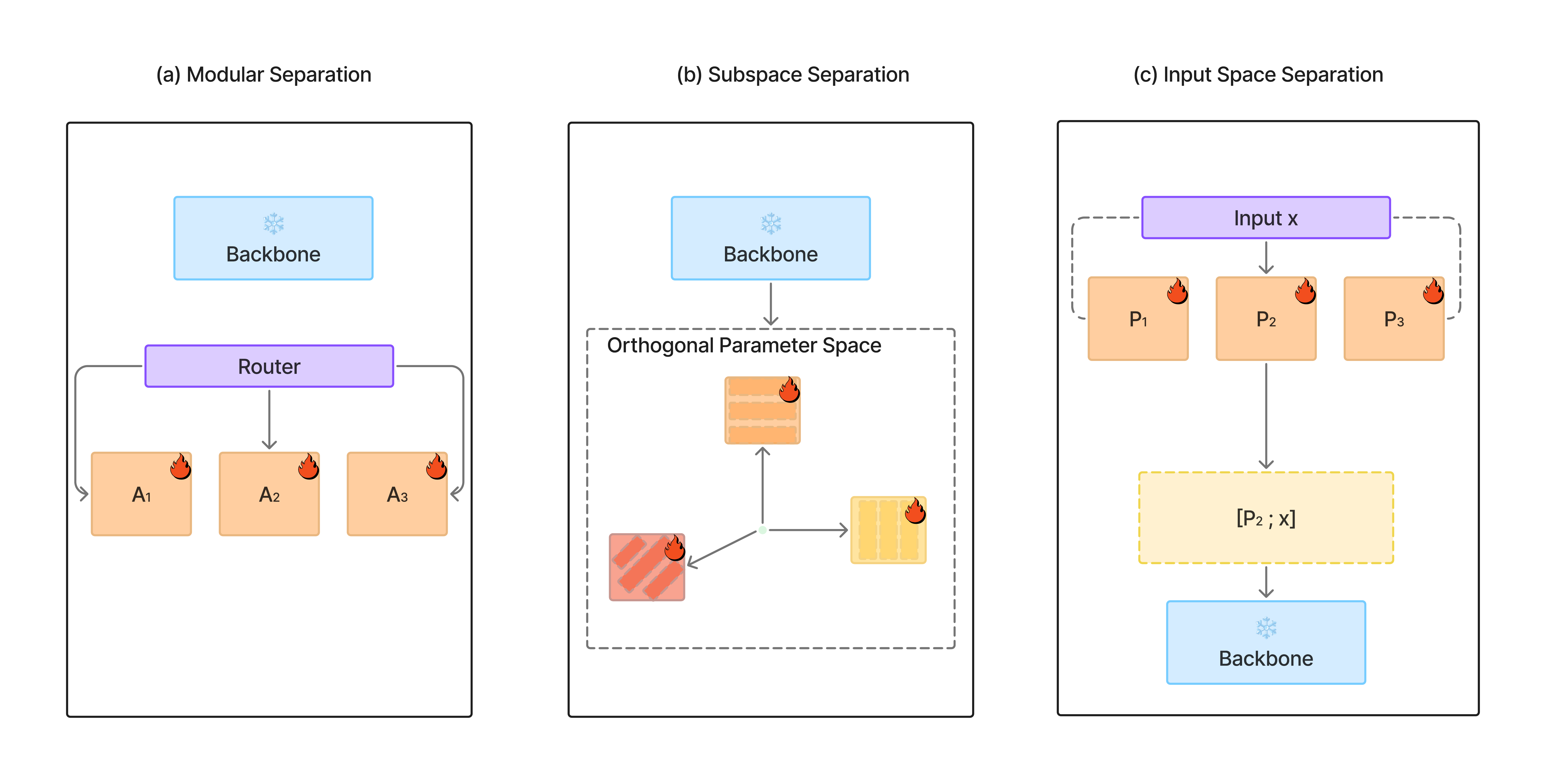}
    \caption{
    The three Parameter Separation strategies, illustrating the main mechanisms by which PECFT methods introduce additional learning capacity on top of a frozen backbone.
    In Fig. 3a, $A_x$ indicates task-specific adapters or modules trained independently;
    Fig 3b illustrates multiple distinct parameter subspaces;
    Fig 3c shows input space separation via leaned prompts ($P_x$) concatenated to the input.
    }
    \label{fig:parameter_separation}
\end{figure}

To illustrate the impact of this design choice, Figure \ref{fig:parameter_separation} provides a visual representation of the architectures that implement different parameter separation strategies.

\subsubsection{Interference Prevention}
In the context of PECFT, the phenomenon of catastrophic forgetting may also be caused by interference among the different tasks.
Therefore, avoiding this kind of issue requires explicit architectural design choices, which we briefly summarize in the following.

\paragraph{Orthogonality Constraints}
    One approach enforces mathematical orthogonality between task-specific updates through gradient projection techniques that constrain new learning to subspaces orthogonal to previous tasks \cite{qiao2024learn}. Recent theoretical work has identified task-level feature orthogonality as a key factor influencing PECFT performance, alongside training sample size and regularization \cite{liu2025parameterefficientfinetuningcontinuallearning}. This provides the strongest theoretical guarantees against interference by ensuring that updates for new tasks cannot directly modify the directions used by previous tasks. Methods like \cite{liang2024inflora,chen2024duallowrankadaptationcontinual} enforce orthogonality between task-specific updates. This provides the strongest theoretical guarantees against interference, explaining why \cite{liang2024inflora} achieves exceptional stability (4.61 ± 0.44\% forgetting). However, orthogonality constraints limit the available parameter space for each task, creating potential scalability concerns as the number of tasks grows. 

\paragraph{Architectural Modularity}
    This aims to prevent interference through physical separation of parameters, where each task uses dedicated modules while the backbone remains frozen. This architectural separation provides interference resistance through construction rather than restraint, since tasks often use entirely different parameters and they cannot interfere with each other. This includes approaches that use task specific adaptation modules with different kinds of learned routing mechanism to select the adapted modules. Methods like \cite{wang2024selfexpansionpretrainedmodelsmixture,10.1007/s11263-024-02218-0} show competitive results, however with performance that degrade when the number of tasks increase. In fact, while forgetting may seem reasonable on smaller tasks, this kind of separation forces modules to become very task specific. In this scenario, more tasks compete for selection, adding complexity to the selection/routing mechanism. Moreover, a wrong selection could lead to catastrophic failure especially when the task is highly challenging. For example, \cite{wang2024selfexpansionpretrainedmodelsmixture} achieve good performance (86.41\% on CIFAR-100) but shows severe scalability issues on fine-grained tasks, especially when increasing the number of tasks from 10 to 20 (CUB-200: 74.19\% → 65.15\%, \textbf{9.04\%} drop).

\paragraph{Compositional Isolation}
      A third approach uses compositional mechanisms to isolate task-specific information within shared parameter spaces. This strategy defines reusable components, such as prompt vectors, that can be combined differently for different tasks through selection schemes or weighted coefficient mechanisms, \textit{i.e.} creating a shared pool of reusable components that can be mixed and matched for different tasks. Instead of learning tasks with completely separate PEFT modules, or forcing the task to use the same modules with some kind of constraints, it creates a set of reusable modules where each task  uses a different combination of these modules.
      Well-designed compositional approaches can achieve stability comparable to orthogonal methods (CODA-Prompt: 4.98\% forgetting) through careful component design that minimizes conflicts between different composition strategies. Compositional methods face fundamental representational limits as task diversity increases. In fact, even sophisticated composition strategies eventually encounter situations where existing components cannot adequately represent new task requirements, leading to universal scalability degradation.


\section{Future Directions}\label{Future Directions}

\subsection{Multi-Modality Pre-trained Models}
Different pre-trained models' influence on CL methods' effectiveness requires deeper analysis.
Research not only indicates variations in performance between models trained from scratch and those utilizing pre-trained architectures but also shows that different pre-trained models benefit differently from existing continual learning approaches \cite{lee2023pre}.
Thus, exploring the selection of the optimal pre-trained model and its best-fitted continual learning strategy given a real-world task can lead to meaningful contributions to guiding industry practice.
Additionally, while a majority of the experiments were tested over visual and natural language modalities, video and audio are domains where the application of continual learning can be valuable to explore as well \cite{niizumi2022byol,luo2020univl}.

A multimodality pre-trained model poses a challenge but also offers opportunities for developing cross-modality strategies for continual learning tasks.
The key idea is that as the models receive additional information from the task, they can develop a more accurate and robust representation, especially when the supplementary guidance comes from large pre-trained models, which are known for their ability to generate robust representations in the first place.
For instance, \cite{ma2024language} indicates that aligning visual features with semantic groups and leveraging semantic relations among categories can boost model robustness against distribution shifts. 
While language guidance has been extensively investigated in Transfer Learning~\cite{socher2013zero} across different vision tasks, its utilization in continual learning has been relatively overlooked. 
With the rise of large pre-trained models in specific domains, exploring potential cross-modality guidance holds promise for further research.

\subsection{{Extending PECFT to 3D Data and Point Cloud Analysis}}
{
The majority of current Parameter-Efficient Continual Fine-Tuning research focuses on 2D Computer Vision (CV) and Natural Language Processing (NLP). However, the critical need for robust intelligent systems in fields like autonomous driving, robotics, and AR/VR demands extending PECFT principles to the domain of 3D data, particularly Point Cloud. Point clouds, being sparse, unordered, and highly sensitive to geometric transformations, present unique challenges for efficient and continual adaptation of large pre-trained models. Successfully updating these 3D models (which often utilize transformer or graph-based architectures) to recognize new classes or adapt to new sensor data without incurring catastrophic forgetting is a crucial research frontier.
Early work demonstrates the feasibility of applying PEFT techniques tailored for 3D data:
}

\begin{itemize}
    \item {Prompt- and Adapter-based Fusion: Methods like DAPT (Dynamic Adapter Meets Prompt Tuning) \cite{zhou2024dynamic} and Point-PEFT \cite{tang2024pointpeftparameterefficientfinetuning3d} combine adapters and prompts, often dynamically generated, to efficiently incorporate downstream 3D semantics while freezing the backbone.}
    
    \item {Spectral Domain Tuning: Approaches such as PointGST (Point cloud Graph Spectral Tuning) \cite{liang2025parameter} introduce lightweight, trainable adapters that fine-tune parameters in the spectral domain, leveraging graph Fourier transforms to capture intrinsic geometric information and mitigate confusion in the spatial domain.}
\end{itemize}

{
These initial efforts confirm that hybrid, parameter-efficient strategies are essential for realizing scalable, adaptable 3D intelligent systems, making the integration of PECFT into this domain a critical direction for future research.
}

\subsection{Model Merging}
Model merging presents an exciting opportunity in  CL, by combining multiple expert models, each specialized in different aspects of a task, we can create a system that not only mitigates issues like catastrophic forgetting but also benefits from the diverse strengths of each model. This becomes particularly important in dynamic, evolving domains, as it allows the model to expand its knowledge over time without forgetting what it has previously learned. However, a common problem that these model merging solutions, such as \textbf{Task Arithmetic }\cite{ilharco2023editingmodelstaskarithmetic}, often encounter is parameter interference, which leads to significant performance degradation when these expert models are merged.
Some works such as \textbf{TIES-Merging} \cite{yadav2023ties} and \textbf{DARE} \cite{yu2024language} have led to significant improvements in model merging. TIES addresses interference by resetting parameters that have only changed minimally, resolving sign conflicts, and merging only those parameters that align with the final agreed-upon sign. DARE, on the other hand, eliminates redundant delta parameters by randomly dropping them and rescaling the remaining ones, which has shown tremendous effectiveness in sparsifying and merging multiple expert models without significant performance loss. 

Typical model merging scenarios often require combining pre-existing expert models, each specialized in a specific task, into one unified system. However, this static approach falls short in scenarios where new tasks emerge over time. In continual learning, we face the challenge of incrementally integrating new task-specific models without retraining the entire system. Recent advances in dynamic model merging address this by tackling issues such as parameter interference, memory efficiency, and sequential integration, enabling systems that adapt more effectively as new tasks are encountered. For instance, \textbf{MagMax} \cite{marczak2024magmaxleveragingmodelmerging} introduces a framework that merges task-specific models using sequential fine-tuning combined with a maximum magnitude weight selection strategy. This approach integrates new information effectively and preserves the integrity of earlier learning to help tackle catastrophic forgetting. In contrast, \textbf{Representation Surgery for Multitask Model Learning} \cite{yang2024representationsurgerymultitaskmodel} addresses a different challenge. Here, the focus is on mitigating the representation bias that emerges when merging models trained on disparate tasks. By inserting a lightweight, task-specific module, dubbed “Surgery”, the method realigns the merged model's internal representations with those of the individual models, thereby enhancing overall performance in multitask scenarios. Instead, \textbf{Adaptive LoRA Merging for Domain Incremental Learning} \cite{coleman2024adaptive} highlights the limitations of fixed-weight merging by proposing an adaptive mechanism that dynamically computes merging coefficients. This flexibility allows the system to balance the contributions of new and old domains, ensuring robust performance in evolving environments while reducing manual tuning.
{
Conversely, \textbf{HAM} \cite{coleman2025hamhierarchicaladaptermerging} adopts a hierarchical merging strategy aimed at mitigating interference among task-specific LoRA modules.
During training, LoRA adapters are grouped based on their similarity, followed by a final merging step that produces a unified model for inference.
This approach has been shown to reduce interference between LoRA adapters and to yield improved performance on longer task sequences.
}
Lastly, \textbf{OPCM} \cite{tang2025mergingmodelsflyretraining} takes a sequential projection-based approach. By projecting new updates onto subspaces orthogonal to those of previously merged models and applying adaptive scaling, this method minimizes interference and maintains a constant memory footprint, making it highly scalable for continual learning applications.
By moving away from traditional continual learning approaches, recent trends focus on the adaptive combination of lightweight modules such as \cite{pmlr-v97-houlsby19a,hu2022lora} in dynamic environments. This enables the seamless integration of new tasks as they emerge, without the need for extensive retraining of large, monolithic models. By merging these modular components on the fly, systems can remain both efficient and realistic in handling real-world challenges, making them ideally suited for large-scale models.

\textcolor{red}{
\subsection{Data Constrained Environments }
Real-world deployment rarely affords the abundant, centrally pooled, and stationary data assumed by most CL benchmarks. Instead, models must adapt under one or more data constraints: novel data may arrive with only a handful of labeled examples and under shifting input distributions, and it may be scattered across many devices or institutions where privacy, bandwidth, or regulation forbid centralization. These constraints make PECFT especially attractive: by freezing the backbone and updating only a small set of prompts, adapters, or low-rank factors, parameter-efficient methods reduce both the supervision needed to adapt and the volume of information that must be communicated or stored. We highlight two prominent settings that exemplify this challenge.
}

\textcolor{red}{
\subsubsection{Few-Shot Domain Incremental Learning FSDIL}
Standard DIL quietly assumes that every new domain arrives with plenty of labeled data. That assumption tends to fail in the very applications where continual adaptation matters most, such as autonomous driving, robotic vision, and clinical imaging, where a new domain (a change in lighting, sensor, or geography) usually comes with only a handful of labels, and where privacy rules forbid keeping the old data around.  FSDIL formalized in PGO-BEn \cite{mukherjee2026textscpgoben} directly tackles this scenario.  The label space stays fixed, the domain shifts across a stream with no domain identifier at test time, each new domain is seen through only a few labeled samples per class, and no exemplars from earlier sessions can be retained. It therefore stacks three problems on top of each other: severe label scarcity, distribution shift, and an exemplar-free privacy constraint. Earlier prompt-pool methods for DIL struggle here, because the per-domain prompt selection they depend on becomes unreliable once data is scarce. PECFT fits this setting well. A frozen vision-language backbone such as CLIP supplies transferable, domain-robust priors, while a small set of trainable prompts learn each shift. PGO-BEn is a concrete example: it keeps both CLIP encoders frozen, learns only lightweight prompts, and conditions the text prompts on the visual prompts across layers so the text embedding space tracks the changing visual distribution and stays domain-agnostic, with no domain routing at inference. To limit forgetting without storing gradient subspaces, it builds a proxy of past-domain knowledge from the current model and pushes any conflicting update into an orthogonal direction. It also swaps the usual exponential moving average for a Beta-function temporal ensemble that weights both early and late training states, which helps it hold on to knowledge from the first domains. Taken together, this multi-modal prompting, gradient-aligned regularization, and parameter-light ensembling show how PECFT can support rehearsal-free adaptation even when data is this constrained.
}

\textcolor{red}{
\subsection{Few-Shot Class Incremental Learning (FSCIL) }
FSCIL requires a model to learn classes  from only a few labeled instances per class while preserving performance on previously learned ones, placing it at the sharp end of the stability-plasticity trade-off \cite{tao2020few}. Under a constrained parameter budget, full-network fine-tuning is poorly suited to this scenario, since the scarcity of new samples induces overfitting and accelerates catastrophic forgetting of previous classes \cite{YUAN2025287}. Recent methods lean instead on architectural or representation-level efficiency. Prompt-based approaches, for instance, insert small trainable domain and task-specific prompts into the layers of a frozen Vision Transformer through prefix-tuning, so that each incremental update stays separated from the features the model has already learned \cite{YUAN2025287}.An alternative approach, UPP \cite{jiang2026unlocking},  freezes the feature extractor outright to protect past representations and shifts the work into a lightweight decision space, learning dynamic class prototypes that can be tuned cheaply as new classes arrive. By limiting updates to targeted adapter modules or to the decision boundary itself, these techniques keep representation drift under control and let a model follow a continuous stream of new classes with very little compute and very little data \cite{han2021pretrained,YUAN2025287}.
}

\textcolor{red}{
\subsection{Decentralized and Cross-Client Parameter Efficient Continual Fine-Tuning }
Most work on parameter-efficient continual fine-tuning (PECFT) so far assumes a single model learning a sequence of tasks over time. A promising but much less explored direction is what happens when that learning is spread across many devices, as in Federated Class-Incremental Learning (FCIL). In this setting the model has to handle change along two axes at once, not only over time but also across clients whose data can look very different from one another. The result is a kind of double forgetting. Each client slowly loses its earlier knowledge as its own data stream evolves, and the central server loses information too, since the updates it gathers from different clients are heterogeneous and non-IID, and averaging them tends to blur what any single client has learned \cite{dong2023no,luo2025federated}.Progress here will depend on adaptation methods that let clients cooperate without sharing raw data, all while keeping the foundation model frozen. One natural idea is to maintain distributed, task-aware pools of prompts that separate the knowledge shared across clients from the parts specific to each one, and to do this without a central memory buffer \cite{luo2025federated}. Another is to rely on orthogonal low-rank updates, such as localized LoRA variants, that guide each client's changes into its own region of the parameter space \cite{guo2024federated}. If those updates can be kept from overlapping when the server merges them, the global model could take on a growing set of new classes without clients interfering with one another. Sorting out this coordination will be central to running lightweight, lifelong learning across large fleets of edge devices.
}

\subsection{From Classification to Reasoning}
Large models have been considered to exhibit ``emergent'' phenomena, yet they still have ways to go before they can effectively handle more complex reasoning tasks \cite{wei2022emergent,schaeffer2024emergent}.
CL is crucial in this regard and will also face new challenges.

The evolution of tasks from single-ability tasks, such as image classification, to more complex reasoning tasks demands a re-evaluation of existing continual learning methods or the proposal of new approaches. 
Classically, many methods in CL are evaluated using classification datasets such as CIFAR and ImageNet. 
There has been a recent push towards transitioning from hard split datasets to benchmarks that reflect more natural temporal and task shifts \cite{lin2021clear,verwimp2023clad}.
Despite this, very few studies address the specific requirement for more complex reasoning tasks, such as visual question answering \cite{marino2021krisp,yi2018neural,lei2023symbolic}, decision-making \cite{dong2019neural}, motion programming \cite{kulal2021hierarchical}, scene understanding \cite{liu2019learning}.

Reasoning tasks can differ significantly in their requirements. For instance, in image classification, a model continually improves its ability to recognize objects within a single modality and primarily focuses on learning representations for each class. However, in reasoning tasks such as visual question answering, a model often needs to acquire multiple new abilities such as attribute recognition, knowledge reasoning based on changing demands \cite{lei2023symbolic,zhang2023vqacl}.

Representation methods that focus on retain representative embedding from historical task typically fail to handle this situation, thus cannot be applied directly.
Traditional replay techniques may prove insufficient in addressing these more intricate tasks as well. 
Some good examples that dive deeper into the specific need for reasoning tasks are: Concept-Level Continual Learning \cite{marconato2023neuro} maps sub-symbolic inputs to high-level concepts and explicitly replay concepts for CL in neuro-symbolic architectures such as DeepProbLog \cite{manhaeve2018deepproblog};
SGP \cite{lei2023symbolic} utilizes scene graph as a prompt to represent previous images in visual question answering task and replays pseudo scene graphs alongside corresponding QA pairs.

Therefore, a promising direction for future research efforts involves developing novel approaches to improve the reasoning ability of large models within the framework of continual learning.

\subsection{More Realistic and Practical Settings}
Future research should explore Continual Learning in more practical and realistic settings.
This includes scenarios characterized by limited computational budgets. A recent survey~\cite{verwimp2023continual} pointed out that existing continual learning methods tend to consider memory constraint, but computational costs are not extensively considered. While the cost of memory and privacy concerns may seem prohibitive, in practice, storing large datasets for extended periods can relatively inexpensive compared to the computational costs of training models on such datasets \cite{prabhu2023online}. This is especially relevant for methods relying on replay mechanisms.

Additionally, investigating Online Learning~\cite{mai2022online,soutif2023comprehensive} paradigms where data arrives in small, incremental batches presents a significant frontier for research in this field. Inherently, the ability to detect incoming data shifts enables a more efficient model update process and can contribute to more robust model performance.

Task-agnostic~\cite{aljundi2019task} and unsupervised~\cite{bagus2022beyond,chen2022semi} are settings that were less explored but have started to gain more attention in recent years. Classic Continual Learning settings often assume availability of data annotation such as class label or task label. However, data labeling efforts can be costly and even not feasible in some real-world industrial situations. Thus, research into unsupervised or Task-agnostic settings can lead to more practical use cases of Continual Learning in real-world scenarios.

\textcolor{red}{
A valuable next step is to move them into dynamic, high-stakes domains such as medical diagnostics, industrial manufacturing, and environmental monitoring, where the data stream is genuinely non-stationary and retraining the full model on every update is often impractical. Tight latency budgets and limited on-device compute already make full retraining a poor fit for deployed systems \cite{muralidhara2025cloraparameterefficientcontinuallearning,zhu-etal-2022-parameter}. Clinical deployment is a natural case. A foundational medical model has to keep absorbing newly identified pathology subtypes or new imaging modalities as it moves between hospital sites, without drifting away from the representations that earlier diagnoses depend on. Rehearsal-free adapter methods point to one way of doing this. Dynamic-rank LoRA, for example, adds task-specific low-rank components and allocates their rank on the fly, mitigating forgetting without storing past data \cite{Bhat2025ParameterEC}. So far these methods have been shown on general vision benchmarks rather than clinical data, so adapting and validating them on medical imaging is itself an open problem.
}

\textcolor{red}{
Industrial manufacturing puts the emphasis on speed and footprint. Models in real-time systems are expected to pick up tasks such as automated quality control, surface anomaly detection, or predictive maintenance as production lines change, sometimes on hardware as small as a micro-controller. This rests on updating as few parameters as possible per task. Existing work in an industrial real-time setting already shows the shape of the solution, selecting a compact parameter subset offline and regularizing it online so that only the changed parameters need to be stored, which avoids both downtime and a steadily growing model \cite{zhu-etal-2022-parameter}. Carrying this from the text-classification system where it was demonstrated to the vision and sensor tasks typical of the factory floor could prove to be useful.
Environmental monitoring stretches adaptation across space and time. Models working with remote-sensing imagery, climate records, or biodiversity telemetry have to track shifting seasonal baselines and adapt to new geographic regions as they appear. Much of remote sensing is a segmentation problem, and recent work shows that low-rank adaptation can drive class-incremental semantic segmentation at a fraction of the usual cost \cite{muralidhara2025cloraparameterefficientcontinuallearning}, which makes it a credible starting point. The harder, open question is how to explicitly separate the slow, stationary structure of the planet from fast-changing local anomalies, for instance through prompt-based mixtures of experts or per-region low-rank adapters that specialize without relearning the global picture.
}

\section{Conclusion}\label{Conclusion}
In this work, we tackled a key limitation of current foundation models: their restricted ability to adapt to specific downstream tasks over time.
To address this, we focused on two research fields aimed at enhancing adaptation in dynamic scenarios: Continual Learning and Parameter-Efficient Fine-Tuning.
On one hand, CL enables models to incrementally learn from a continuous stream of tasks, while PEFT facilitates rapid adaptation to individual tasks.
We explored the intersection of these two fields , which we called Parameter-Efficient Continual Fine-Tuning (PECFT), and examined how their techniques can be combined to improve the efficiency and sustainability of large pre-trained models in evolving environments.
We believe that bridging CL and PEFT has the potential to drive significant advancements in AI research, shaping the future of large-scale models.
Through this review, we aim to provide researchers with an overview of existing methods in this emerging area, fostering the development of novel and effective solutions.

\section*{Declaration of Generative AI and AI-assisted technologies in the writing process}
During the preparation of this work the authors used ChatGPT in order to enhance readability and check grammar.
After using this tool, the authors reviewed and edited the content as needed and take full responsibility for the content of the published article.

\section*{Acknowledgments}
Research partly funded by PNRR - M4C2 - Investimento 1.3, Partenariato Esteso PE00000013 - ``FAIR - Future Artificial Intelligence Research'' - Spoke 1 ``Human-centered AI'', funded by the European Commission under the NextGeneration EU programme and Leonardo Labs.
This research was partially supported by the FIS2 Grant from the Italian Ministry of University and Research (MUR), Grant No. FIS2023-03382, under the project “Continual, Decentralized Compositionality for Sustainable Artificial Intelligence” with Prof. Vincenzo Lomonaco serving as Principal Investigator (PI).

\bibliographystyle{spmpsci}
\bibliography{references}

\end{document}